\documentclass[journal]{IEEEtran}
\usepackage{booktabs}
\usepackage{url}
\usepackage{epsfig}
\usepackage{subfigure}
\usepackage{graphics}
\usepackage{graphicx}
\usepackage{formular}
\usepackage{algorithm}
\usepackage{algorithmic}
\usepackage{multirow}
\usepackage{multicol}
\usepackage{xcolor}
\usepackage{clrscode}
\usepackage{makecell}
\usepackage{CJK}
\usepackage{amsthm}
\usepackage{balance}
\usepackage{times}
\usepackage{amsmath}
\usepackage{amsfonts}
\usepackage{stfloats}
\usepackage{array}
\usepackage{float}
\usepackage{url}
\usepackage{epstopdf}

\begin{document}
\title{Unsupervised Feature Learning Architecture with Multi-clustering Integration RBM}
\author{
\IEEEauthorblockN{ Jielei Chu,~\IEEEmembership{member,~IEEE},  Hongjun Wang, Jing Liu,\\
Zhiguo Gong,~\IEEEmembership{Senior member,~IEEE}, Tianrui Li,~\IEEEmembership{Senior member,~IEEE}}
\thanks{Jielei Chu, Hongjun Wang, Tianrui Li (the corresponding author) are with the Institute of Artificial Intelligence, School of Information Science and Technology, Southwest Jiaotong University, Chengdu 611756, China. Tianrui Li is also with National Engineering Laboratory of Integrated Transportation Big Data Application Technology, Southwest Jiaotong University, Chengdu 611756, China. e-mails: \{jieleichu, wanghongjun, trli\}@swjtu.edu.cn.}
\thanks{Jing Liu is with the School of Business, Sichuan University, Sichuan, 610065, Chengdu, China. e-mail: liujing@scu.edu.cn.}
\thanks{Zhiguo Gong is with the State Key Laboratory of Internet of Things for Smart City, Department of Computer and Information Science, University of Macau, Macau, China. Email: fstzgg@um.edu.mo}
}
\maketitle
\begin{abstract}
In this paper, we present a novel unsupervised feature learning architecture, which consists of a multi-clustering integration module and a variant of RBM termed multi-clustering integration RBM (MIRBM). In the multi-clustering integration module, we apply three unsupervised K-means, affinity propagation and spectral clustering algorithms to obtain three different clustering partitions (CPs) without any background knowledge or label. Then, an unanimous voting strategy is used to generate a local clustering partition (LCP). The novel MIRBM model is a core feature encoding part of the proposed unsupervised feature learning architecture. The novelty of it is that the LCP as an unsupervised guidance is integrated into one step contrastive divergence ($\textbf{\texttt{\texttt{CD}}}_{1}$) learning to guide the distribution of the hidden layer features. For the instance in the same LCP cluster, the hidden and reconstructed hidden layer features of the MIRBM model in the proposed architecture tend to constrict together in the training process. Meanwhile, each LCP center tends to disperse from each other as much as possible in the hidden and reconstructed hidden layer during training. The experiments demonstrate that the proposed unsupervised feature learning architecture has more powerful feature representation and generalization capability than the state-of-the-art graph regularized RBM (GraphRBM) for clustering tasks in the Microsoft Research Asia Multimedia (MSRA-MM)2.0 dataset.
\end{abstract}
\begin{IEEEkeywords}
   multi-clustering integration RBM; unsupervised feature learning; $\textbf{\texttt{\texttt{CD}}}_{1}$ learning; image clustering.
\end{IEEEkeywords}
\IEEEpeerreviewmaketitle
\section{Introduction}
Feature learning is a crucial phase in many applications (e.g. visual recognition \cite{Yuan2017Learning}, scene analysis \cite{Gao2016A}, object recognition \cite{Diao2015Object}, multimodal learning \cite{choo2018learning}, \cite{Li2017Multi}, speech recognition \cite{Ghahabi2018Restricted}, image classification \cite{8478774}). Supervised feature learning has achieved great success in machine learning \cite{Lecun2015Deep}. However, the labeled data is scarce in many applications. In recent years, some works focus on the unsupervised feature learning method \cite{Chen2017Graph, 6807781, 6990594, 7097079, 7155531, 8082527}. But, how to obtain appropriate features distribution without any background is still a hard problem in machine learning. To address this, a novel unsupervised feature learning architecture with multi-clustering integration RBM is designed in this paper.\\
\indent Many modeling paradigms such as autoencoders and energy-based models have been applied to feature learning. The restricted Boltzmann machine (RBM) \cite{hinton1986learning} is a popular energy-based model for unsupervised feature learning and aims to explore appropriate hidden features. The structure of a RBM is a bipartite graph consisting of a binary visible layer and a binary hidden layer. There are no connections between the visible layer units and the hidden layer units. The most popular learning algorithms of RBM such as stochastic maximum likelihood \cite{Tieleman2008Training} and contrastive divergence (CD) \cite{hinton2002training} base on the efficient Gibbs sampling. There are a large number of successful applications based on the RBMs, e.g., speaker recognition \cite{Ghahabi2018Restricted}, feature fusion \cite{Li2017Multi}, clustering \cite{Chen2017Graph}, classification \cite{Lu2016A}, \cite{finkfuzzy2015fuzzyclassification}, \cite{chenspectral2015SpatialClassificationDBF}, computer vision \cite{nie2015generative} and speech recognition \cite{graves2013speech}. Meanwhile, various variants of the RBMs have been proposed by the researchers, e.g., pairwise constraints RBM with Gaussian visible units (pcGRBM) \cite{Chujielei2018pcGRBM}, classification RBM \cite{yu2014renyi}, fuzzy RBM (FRBM) \cite{PChen2015fuzzyrbm} and spike-and-slab RBM (ssRBM) \cite{courville2014spike}. For real-valued data, the RBM with Gaussian visible units \cite{choo2018learning}, \cite{Chujielei2018pcGRBM} as the canonical energy model has usually been applied to extract the hidden features from image data. Unlike standard RBM, the visible layer units of the model have Gaussian noise and the hidden layer still maintains binary units. The CD learning can also be used to train the RBM with Gaussian visible units \cite{karakida2016dynamical}. The hidden representations of traditional RBMs do not have explicit instance-level constraints. So, Chu et al. presented semi-supervised pcGRBM in which pairwise constraints are fused into the reconstructed visible layer \cite{Chujielei2018pcGRBM}. However, the labeled data is lacking in many applications and it is expensive to obtain more labels. So, it is certainly worth exploring unsupervised feature learning method of RBMs by fusing external interventions. \\
\indent In this paper, we present a novel unsupervised feature learning architecture, which consists of a multi-clustering integration module and a variant of RBM termed multi-clustering integration RBM (MIRBM). In the multi-clustering integration module, we choose three unsupervised K-means, affinity propagation (AP) and spectral clustering (SC) algorithms to obtain three different global clustering partitions (CPs). Then, an unanimous voting strategy is used to generate the local clustering partition (LCP) of visible layer data. Hence, the LCP only has partial visible layer data. The visual example is shown in Fig. 2. The novel MIRBM model is a core feature encoding part of the proposed unsupervised feature learning architecture. The novelty of it is that the LCP as an unsupervised guidance is integrated into the $\textbf{\texttt{\texttt{CD}}}_{1}$ learning to guide the distribution of the hidden layer features. For the instance in the same LCP cluster, the hidden and reconstructed hidden layer features of the MIRBM model in the proposed architecture tend to constrict together in the training process. Meanwhile, each LCP center tends to disperse from each other as much as possible in the hidden and reconstructed hidden layer during training. As far as we know, this is the first work to use the LCP as as an unsupervised guidance to guide the distribution of the hidden layer features of the proposed MIRBM model in the unsupervised feature learning architecture. The contributions of our work are summarized below.
\begin{itemize}
  \item A novel unsupervised feature learning architecture is proposed, which consists of a multi-clustering integration module and an MIRBM model. The multi-clustering integration module is used to obtain unsupervised guidance information. The MIRBM model is a core feature encoding part to extract hidden layer features.
  \item In the multi-clustering integration module of the proposed architecture, three unsupervised algorithms are employed to obtain three different global CPs without any background knowledge or label. The LCP is generated by the unanimous voting strategy from three different CPs.
  \item The MIRBM model in the proposed architecture uses the LCP as an unsupervised guidance to guide the distribution of the hidden layer features by integrating the LCP into the $\textbf{\texttt{\texttt{CD}}}_{1}$ learning. For the instance in the same LCP cluster, the hidden and reconstructed hidden layer features of the MIRBM modle in the proposed architecture tend to constrict together in the training process. Meanwhile, each LCP center tends to disperse from each other as much as possible in the hidden and reconstructed hidden layer during training.
  \item It is demonstrated that the proposed architecture has more powerful feature representation and generalization capability than the state-of-the-art GraphRBM in the Microsoft Research Asia Multimedia (MSRA-MM)2.0 dataset for image clustering task.
\end{itemize}
\indent The remaining of the paper is organized as follows. The literature review is provided in Section \uppercase\expandafter{\romannumeral2}. In Section \uppercase\expandafter{\romannumeral3}, the theoretical background is described. An unsupervised feature learning architecture together with the MIRBM model are proposed in Section \uppercase\expandafter{\romannumeral4}. The experimental results are shown in Section \uppercase\expandafter{\romannumeral5}. Finally, we conclude in Section \uppercase\expandafter{\romannumeral6} and discuss a few correlative future work.
\section{Literature Review}
 In this section, we review literature on supervised, semi-supervised, unsupervised feature learning based on RBMs and other models, together with the voting strategy in supervised learning.\\
\indent Supervised feature learning has proved to be an effective method in machine learning \cite{8478774}, \cite{Amer2017Deep}, \cite{Lecun2015Deep}, \cite{Luo2018Enhancing}, \cite{Alani2017Arabic}. Amer et al. \cite{Amer2017Deep} proposed a Multimodal Discriminative CRBMs (MMDCRBMs) model based on a Conditional RBMs (an extension of the RBM). Its training process is composed of training each modality using labeled data and training a fusion layer. For multi-modality deep learning, Bu et al. \cite{Bu20173D} developed a supervised 3D feature learning framework in which a RBM is used to mine the deep correlations of different modalities. Cheng et al. \cite{Cheng2017Duplex} presented a novel duplex metric learning (DML) framework for feature learning and image classification. The main task of DML is to learn an effective hidden layer feature of a discriminative stacked autoencoder (DSAE). In the feature space of the DSAE, similar and dissimilar samples are mapped close to each other and further apart, respectively. This framework is the most related work to our study, but it belongs to supervised feature learning with a DSAE by layer-wisely imposing metric learning method and it is applied to image classification tasks.\\
\indent However, supervision information, e.g., labels, is scarce and it is expensive to obtain more labels in many applications. So, some works \cite{chengang2015deep}, \cite{chen2017rolling}, \cite{Chujielei2018pcGRBM} explored semi-supervised feature learning which only needs a small number of labels. Chu et al. \cite{Chujielei2018pcGRBM} presented a pcGRBM model by fusing pairwise constraints into the reconstructed visible layer for clustering tasks. To mitigate the burden of annotation, Yesilbek and Sezgin \cite{yesilbek2017sketch} applied self-learning methods to build a system that can learn from large amounts of unlabeled data and few labeled examples for sketch recognition. The systems perform self-learning by extending a small labeled set with new examples which are extracted from unlabeled sketches. Chen et al. \cite{chen2017rolling} developed a deep sparase auto-encoder network with supervised fine-tuning and unsupervised layer-wise self-learning for fault identification.\\
\indent The class RBMs have powerful unsupervised feature learning capability. To exploit more powerful feature leaning ability, many unsupervised feature learning approaches based on the RBMs have been proposed by previous researches \cite{8520760}, \cite{Keyvanrad2017Effective}, \cite{Tang2016Hidden}, \cite{Chopra2018Restricted}, \cite{Zhang2017Recursive}, \cite{Chen2017Graph}, \cite{Xie2016Finding}. Chopra and Yadav \cite{Chopra2018Restricted} presented a unique technique to extract fault feature from the noisy acoustic signal by an unsupervised RBM. Zhang et al. \cite{Zhang2017Recursive} proposed unsupervised feature learning based on recursive autoencoders network (RAE) for image classfication. They used the spectral and spatial information from original data to produce high-level features. Xie et al. \cite{Xie2016Finding} showed a novel approach to optimize RBM pre-training by capturing principal component directions of the input with principal component analysis. Al-Dmour and Al-Ani \cite{al2018clustering} proposed a fully-automatic segmentation algorithm in which a neural network (NN) model is used to extract the features of the brain tissue image and is trained using clustering labels produced by three clustering algorithms. The obtained classes are combined by majority voting. The study is closely related to our work, but our encoding framework based on RBMs is guided by self-learning local supervisions which stem from unsupervised clustering algorithms and unanimous voting strategy. More specifically, these self-learning local supervisions from visible layers are integrated into the CD learning of RBMs to constrict and disperse the distribution of the hidden layer features and reconstructed hidden layer features. Stewart and Ermon \cite{stewart2017label} presented a new technique to supervise NN by prior domain knowledge for computer vision tasks. It is a related work to our study. However, their work faces to a convolutional neural network (CNN) and requires large amounts of prior domain knowledge and how to encode prior knowledge into loss functions of a CNN is a new challenge.\\
\indent Two existing voting strategies are often used to supervised learning in previous researches. One is the max-voting scheme. For example, Azimi et al. \cite{Azimi2017Advanced} developed a deep learning method for low carbon steel microstructural classification via fully CNN (FCNN) accompanied by a max-voting scheme. The other is the majority voting scheme. For example, Seera et al. \cite{Seera2017Classification} applied a recurrent NN (RNN) to extract features from the Transcranial Doppler (TCD) signals for classification tasks. This work proposed an ensemble RNN model in which the majority voting scheme is used to combine the single RNN predictions. Recently various voting classifiers using majority voting have been proposed to enhance the performance of the classification \cite{Verma2016Comparison}, \cite{Abdalmalak2016Enhancement}, \cite{Guan2017Classifying}, \cite{Tahir2018Efficient}, \cite{Hosseini2018Random}, \cite{He2018}.\\
\indent Chen et al. \cite{Chen2017Graph} illustrated a new graph regularized RBM (GraphRBM) to extract hidden layer representations for unsupervised clustering and classification problem. Meanwhile, they have considered the manifold structure of the image data. The GraphRBM is a state-of-the-art unsupervised feature learning model and it is also the most relevant work to our method. Hence, we compare our unsupervised feature learning architecture with the GraphRBM in the experiments.
\section{Theoretical Background}
\subsection{Restricted Boltzmann Machine}
  A RBM \cite{hinton1986learning} consists of two-layer structure: a visible layer and a hidden layer with stochastic binary units via symmetrically weighted connections. It has no interior-layer connections both between the visible layer units and between the hidden layer units. An energy function of a joint distribution of the visible layer and hidden layer units takes the form:
   \begin{equation}
   E(\textbf{v},\textbf{h})=-\sum\limits_{i\in visibles}a_{i}v_{i}-\sum\limits_{j\in hiddens}b_{j}h_{j}-\sum\limits_{i,j}v_{i}h_{j}w_{ij},
   \end{equation}
  where $\textbf{v}$ is the visible layer vector and $v_{i}$ is the binary states of visible unit $i$, $\textbf{h}$ is the hidden layer vector and $h_{j}$ is the binary states of hidden unit $j$, $a_{i}$ and $b_{j}$ are the biases of visible layer and hidden layer respectively, $w_{ij}$ is the symmetric connection weight between $v_{i}$ and $h_{j}$. \\
  \indent The probability distribution over a vector $\mathbf{v}$ and with the parameters $\theta=\{\textbf{a},\textbf{b},\textbf{W}\}$ takes the form:
   \begin{equation}
   p(\textbf{v},\theta)=\frac{e^{-E(\textbf{v},\theta)}}{Z(\theta)}
   \end{equation}
where $Z(\theta)=\sum\limits_{\textbf{v}}e^{-E(\textbf{v},\theta)}$ is a normalisation constant, $\textbf{a}$ is a vector of the visible layer biases, $\textbf{b}$ is a vector of the hidden layer biases, $\textbf{W}$ is connection matrix. The conditional probability distributions of hidden layer and visible layer units of the RBM are given by:
 \begin{equation}
   p(h_{j}=1|\textbf{v})=\sigma(b_{j}+\sum\limits_{i}v_{i}w_{ij})
\end{equation}
and
\begin{equation}
   p(v_{i}=1|\textbf{h})=\sigma(a_{i}+\sum\limits_{j}h_{j}w_{ij}),
\end{equation}
where $\sigma$ is the sigmoid function.
\subsection{Gaussian Linear Visible Units}
The classical RBM was designed with binary units for both the hidden and visible layers \cite{hinton2002training}. For training real-valued data, the visible layer of RBM consists of Gaussian linear units and the hidden layer of RBM is still binary units. The energy function of RBM with Gaussian linear visible units takes the form:
   \begin{equation}
   E(\textbf{v},\textbf{h})=-\sum\limits_{i\in visibles}\frac{{(v_{i}-a_{i})}^2}{2\sigma_{i}^2}-\sum\limits_{j\in hiddens}b_{j}h_{j}-\sum\limits_{i,j}\frac{v_{i}}{\sigma_{i}}h_{j}w_{ij},
   \end{equation}
  where $\sigma_{i}$ is the standard deviation of visible unit $i$ with Gaussian noise. In the visible layer, the conditional probability is defined by:
\begin{equation}
\begin{aligned}
  p(\textbf{v}|\textbf{h})=\mathcal{N}(\sum\textbf{h}\textbf{W}^T+,\sigma^{2}),
\end{aligned}
 \end{equation}
 where $\mathcal{N}(\cdot)$ represents gaussian density ($\mu=\sum\textbf{h}\textbf{W}^T+\textbf{a}$). The update rules of the parameters become simple when the linear visible units have unit variance of Gaussian noise. Then, the reconstructed values of Gaussian linear visible units are equal to their top-down input values from the binary hidden units plus their bias.
 \subsection{Contrastive Divergence Learning}
 To learn the parameters of symmetric connection weight of the RBM by Maximum-likelihood (ML) learning \cite{hinton2002training}, the update rule is given by:
 \begin{equation}
 \Delta w_{ij} =\varepsilon(<v_{i}h_{j}>_{0}-<v_{i}h_{j}>_{\infty}),
\end{equation}
where $\varepsilon$ is a learning rate, the angle brackets denote the expectations of the distribution, $<v_{i}h_{j}>_{\infty}$ represents the expectations under the distribution of the RBM model. But it is very hard to obtain unbiased sample of $<v_{i}h_{j}>_{\infty}$.\\
\indent So, a faster learning algorithm \cite{hinton2002training} was proposed by applying approximation of the gradient of CD. Karakida et al. \cite{karakida2016dynamical} demonstrated that $\textbf{\texttt{\texttt{CD}}}_{1}$ learning is simpler than  Maximum-likelihood (ML) learning in RBMs. The $\textbf{\texttt{CD}}_{1}$ learning follows the gradient of the difference of two divergences approximately as follows:
\begin{equation}
\begin{aligned}
   \texttt{CD}_{1}=\texttt{KL}(p_{0}||p_{\infty})-\texttt{KL}(p_{1}||p_{\infty}),
 \end{aligned}
\end{equation}
where $\texttt{KL}(p_{0}||p_{\infty})=\sum \limits_{\textbf{v}}p_{0}(\textbf{v})\texttt{log}\frac{p_{0}(\textbf{v})}{p(\textbf{v};\theta})$ and $\texttt{KL}(p_{1}||p_{\infty})=\sum \limits_{\textbf{v}}p_{1}(\textbf{v})\texttt{log}\frac{p_{1}(\textbf{v})}{p(\textbf{v};\theta)}$ are Kullback-Leibler divergences, $p_{0}$ is the distribution of the data, $p_{1}$ is the first step distribution of the Markov chain and $p_{\infty}$ is the distribution of the model. Then the change of symmetric connection weight with $\textbf{\texttt{\texttt{CD}}}_{1}$ learning is given by:
\begin{equation}
 \Delta w_{ij} =\varepsilon(<v_{i}h_{j}>_{0}-<v_{i}h_{j}>_{1}),
\end{equation}
 where the hidden layer units are driven by visible data, $<v_{i}h_{j}>_{0}$ denotes the expectations under the distribution of visible and hidden layer units, $<v_{i}h_{j}>_{1}$ represents the expectations under the distribution of reconstructed visible and hidden layer units. Similarly, the changes of biases $a_{i}$ and $b_{j}$ with $\textbf{\texttt{\texttt{CD}}}_{1}$ learning are given by:
 \begin{equation}
 \Delta a_{i}=\varepsilon(<v_{i}>_{0}-<v_{i}>_{1})
\end{equation}
and
\begin{equation}
 \Delta b_{j}=\varepsilon(<h_{j}>_{0}-<h_{j}>_{1}).
\end{equation}
\indent So, the update rules of all parameters take the form
 \begin{equation}
w_{ij}^{(\tau+1)}=w_{ij}^{(\tau)}+\varepsilon(<v_{i}h_{j}>_{0}-<v_{i}h_{j}>_{1}),
 \end{equation}
 \begin{equation}
 a_{i}^{(\tau+1)}=a_{i}^{(\tau)}+\varepsilon(<v_{i}>_{0}-<v_{i}>_{1})
\end{equation}
and
\begin{equation}
 b_{j}^{(\tau+1)}=b_{j}^{(\tau)}+\varepsilon(<h_{j}>_{0}-<h_{j}>_{1}).
\end{equation}
 The learning efficiency can be obviously improved by the $\textbf{\texttt{\texttt{CD}}}_{1}$ learning.
\section{Unsupervised Feature Learning Architecture with MIRBM}
In this section, we present a novel unsupervised feature learning architecture. The MIRBM model as a core part of the architecture is used to extract hidden layer features. Then, the detailed inferences for the update rules of the MIRBM model parameters are listed. We design a learning algorithm of the MIRBM model and analyse its convergence and complexity.
\subsection{Architecture}
To explore more effective unsupervised feature learning method, we construct a novel feature learning architecture, which consists of a multi-clustering integration module and an MIRBM model. It is shown in Fig. 1. The multi-clustering integration module is used to generate an unsupervised guidance LCP for the next MIRBM model. Three unsupervised K-means, AP and SC algorithms are applied to obtain three CPs, then an unanimous voting strategy is used to generate the LCP. An example of multi-clustering integration module is shown in Fig. 2.\\
\indent The MIRBM model is a most important part of the architecture. The unsupervised LCP information is integrated into the $\textbf{\texttt{\texttt{CD}}}_{1}$ learning to guide the train process of it. With the LCP assistance, the hidden and reconstructed hidden layer features of MIRBM model tend to constrict together in the training process for the instance in the same LCP cluster. Meanwhile, each LCP center tends to disperse from each other as much as possible in the hidden and reconstructed hidden layer during training.
\begin{figure*}
\vspace{1mm} \centering
\includegraphics[scale=0.48]{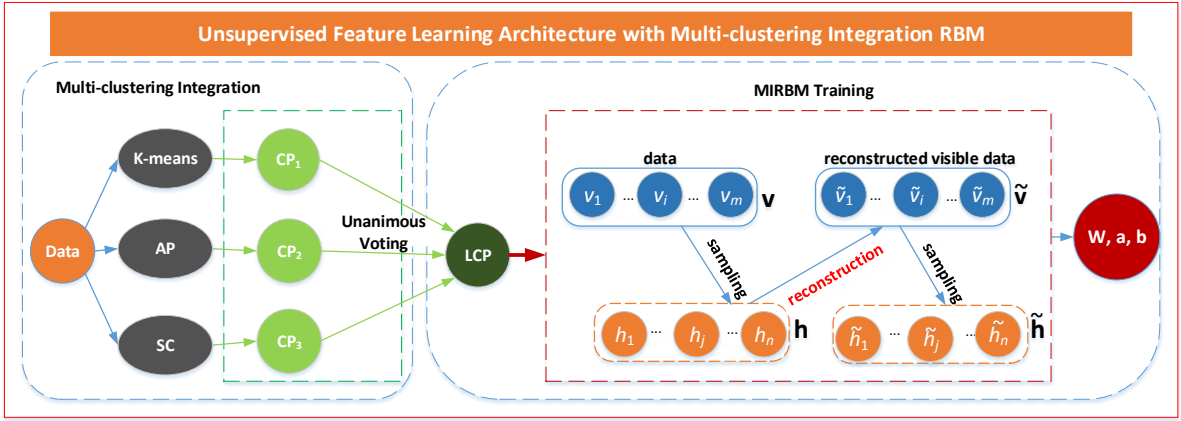}
\caption{Unsupervised feature learning architecture with MIRBM model. The multi-clustering integration module is used to obtain unsupervised guidance information LCP by three unsupervised clustering algorithms and unanimous voting strategy. The MIRBM model as a core feature encoding part uses unsupervised LCP to guide the distribution of the hidden layer features by integrating the LCP into its training process of $\textbf{\texttt{\texttt{CD}}}_{1}$ learning.
} \label{fig:1}
\end{figure*}
\begin{figure*}
\vspace{1mm} \centering
\includegraphics[scale=0.4]{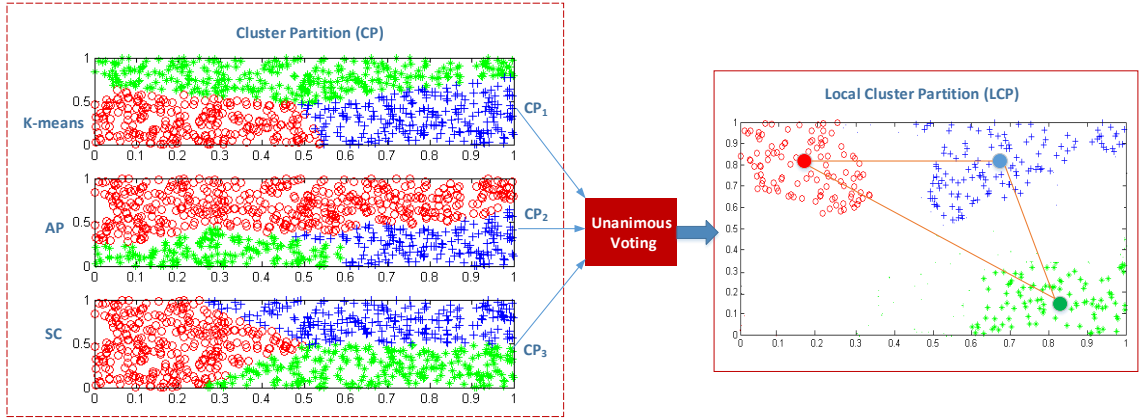}
\caption{An example of multi-clustering integration module. Left: three kinds of different CPs which contain all instances are generated by K-means, AP and SC algorithms. Right: an unanimous voting strategy is used to produce the LCP with three kinds of CPs. The LCP only contains a portion of the instance and three dots represent three local centers of it.
} \label{fig:1}
\end{figure*}
\subsection{The MIRBM model}
\begin{table}
\begin{center}
\caption{List of symbols.}
 \label{imagedata}
\scalebox{1}{
\begin{tabular}{lcccc}
\toprule[1.5pt] 
{Notation}  &  {Definition} \\
\hline
{$\texttt{V}_{data}$} &{Visible layer data set} \\
{$\texttt{H}_{data}$} &{Hidden layer feature set} \\
{$\texttt{V}_{recon}$}& {Reconstructed visible layer set} \\
{$\texttt{H}_{recon}$}& {Hidden layer feature of reconstructed visible layer set} \\
{$\mathbf{\textbf{v}}_{s}, \mathbf{\textbf{v}}_{t}$} &{Visible layer row vector} \\
{$\mathbf{\textbf{h}}_{s}, \mathbf{\textbf{h}}_{t}$} &{Hidden layer feature row vector}\\
{$\mathbf{\widetilde{\textbf{v}}}_{s}, \mathbf{\widetilde{\textbf{v}}}_{t}$} &{Reconstructed visible layer row vector} \\
{$\mathbf{\widetilde{\textbf{h}}}_{s}, \mathbf{\widetilde{\textbf{h}}}_{t}$} &{Hidden layer feature row vector of reconstructed data}\\
{$\texttt{V}_{k}$} &{All vectors of $\texttt{V}_{k}\subset \texttt{V}_{data}$ belonging to the same cluster.} \\
{$\widetilde{\texttt{V}}_{k}$} &{All vectors of $\widetilde{\texttt{V}}_{k}\subset \texttt{V}_{recon}$ belonging to the same cluster.} \\
{$\texttt{H}_{k}$} &{All vectors of $\texttt{H}_{k}\subset\texttt{H}_{data}$ belonging to the same cluster.} \\
{$\widetilde{\texttt{H}}_{k}$} &{All vectors of $\widetilde{\texttt{H}}_{k}\subset\texttt{H}_{recon}$ belonging to the same cluster.} \\
{$\textbf{C}_{k}$} &{The center of cluster $\texttt{H}_{k}$ } \\
{$\widetilde{\textbf{C}}_{k}$} &{ The center of cluster $\widetilde{\texttt{H}}_{k}$} \\
{$\textbf{O}_{k}$} &{The center of cluster $\texttt{V}_{k}$} \\
{$\widetilde{\textbf{O}}_{k}$} &{ The center of cluster $\widetilde{\texttt{V}}_{k}$} \\
\bottomrule[1pt] 
\end{tabular}}
\end{center}
\end{table}
\indent Suppose that $\texttt{V}_{data}=\{\mathbf{\textbf{v}_{1}},\mathbf{\textbf{v}_{2}},\cdots,\mathbf{\textbf{v}_{N}}\}$ is the original data set. $\texttt{H}_{data}=\{\mathbf{\textbf{h}_{1}},\mathbf{\textbf{h}_{2}},\cdots,\mathbf{\textbf{h}_{N}}\}$ is the hidden layer feature set. $\texttt{V}_{recon}=\{\mathbf{\widetilde{\textbf{v}}_{1}}, \mathbf{\widetilde{\textbf{v}}_{2}}, \cdots, \mathbf{\widetilde{\textbf{v}}_{N}}\}$ is the reconstructed visible layer data set. $\texttt{H}_{recon}=\{\mathbf{\widetilde{\textbf{h}}_{1}}, \mathbf{\widetilde{\textbf{h}}_{2}}, \cdots, \mathbf{\widetilde{\textbf{h}}_{N}}\}$ is the hidden features set of reconstructed data.
Let $\texttt{V}_{1}, \texttt{V}_{2}, \cdots \texttt{V}_{K}$ be $K$ local clusters of visible layer set $\texttt{V}_{data}$, $\texttt{H}_{i} (i=1,2, \cdots, K)$ are local clusters mapped of $\texttt{V}_{i} (i=1,2, \cdots, K)$, respectively. Similarly, $\widetilde{\texttt{V}}_{1}, \widetilde{\texttt{V}}_{2}, \cdots, \widetilde{\texttt{V}}_{K}$ are $K$ local clusters of reconstructed visible layer set $\texttt{V}_{recon}$, $\widetilde{\texttt{H}}_{i}(i=1,2, \cdots, K)$ are $K$ local clusters mapped of $\widetilde{\texttt{V}}_{i} (i=1,2, \cdots, K)$, respectively.
We use the gradient descent method to obtain approximate optimal parameters of the MIRBM model. In the encoding process, we expect that the hidden features and the reconstructed hidden features of the instance in the same LCP cluster become more concentrated together. Meanwhile, each LCP center tends to disperse from each other as much as possible in the hidden and reconstructed hidden layer during training of $\textbf{\texttt{CD}}_{1}$ learning method. Therefore, the objective function takes form:
\begin{equation}
\begin{aligned}
   &F(\theta,\texttt{V}_{data})=-\eta\left[\texttt{KL}(p_{0}||p_{\infty})-\texttt{KL}(p_{1}||p_{\infty})\right]+\\
&\Big(\frac{1-\eta}{N_{h}}\sum\limits_{k=1}^{K}\sum\limits_{\mathbf{\textbf{h}}_{s},\mathbf{\textbf{h}}_{t}\in\texttt{H}_{k}}\Arrowvert \mathbf{\textbf{h}}_{s}-\mathbf{\textbf{h}}_{t}\Arrowvert^2-\frac{1-\eta}{N_{C}}\sum\limits_{p=1}^{K-1}\sum\limits_{q=p+1}^{K}\Arrowvert \mathbf{\textbf{C}}_{p}-\\
&\mathbf{\textbf{C}}_{q}\Arrowvert^2\Big)+\Big(\frac{1-\eta}{N_{h}}\sum\limits_{k=1}^{K}\sum\limits_{\widetilde{\mathbf{\textbf{h}}}_{s},\widetilde{\mathbf{\textbf{h}}}_{t}\in\widetilde{\texttt{H}}_{k}}\Arrowvert \widetilde{\mathbf{\textbf{h}}}_{s}-\widetilde{\mathbf{\textbf{h}}}_{t}\Arrowvert^2-\frac{1-\eta}{N_{C}}\sum\limits_{p=1}^{K-1}\sum\limits_{q=p+1}^{K}\Arrowvert \widetilde{\mathbf{\textbf{C}}}_{p}\\
&-\widetilde{\mathbf{\textbf{C}}}_{q}\Arrowvert^2\Big),
\end{aligned}
\end{equation}
where $\eta\in(0,1)$ is a scale coefficient, $N_{h}$ is the cardinality of $\texttt{H}_{k} (k=1,2,\cdots,K)$, $N_{C}=\frac{K(K-1)}{2}$ is the number of pairwise cluster center.\\
\indent To simplify the expression of the objective function, $L_{data}(\theta)$ and $L_{recon}(\theta)$ functions take form:
\begin{equation}
\begin{aligned}
   &L_{data}(\theta)=\\
   &\frac{1}{N_{h}}\sum\limits_{k=1}^{K}\sum\limits_{\mathbf{\textbf{h}}_{s},\mathbf{\textbf{h}}_{t}\in\texttt{H}_{k}}\Arrowvert \mathbf{\textbf{h}}_{s}-\mathbf{\textbf{h}}_{t}\Arrowvert^2-\frac{1}{N_{C}}\sum\limits_{p=1}^{K-1}\sum\limits_{q=p+1}^{K}\Arrowvert \mathbf{\textbf{C}}_{p}-\mathbf{\textbf{C}}_{q}\Arrowvert^2
\end{aligned}
\end{equation}
and
\begin{equation}
\begin{aligned}
   &L_{recon}(\theta)=\\
   &\frac{1}{N_{h}}\sum\limits_{k=1}^{K}\sum\limits_{\widetilde{\mathbf{\textbf{h}}}_{s},\widetilde{\mathbf{\textbf{h}}}_{t}\in\widetilde{\texttt{H}}_{k}}\Arrowvert \widetilde{\mathbf{\textbf{h}}}_{s}-\widetilde{\mathbf{\textbf{h}}}_{t}\Arrowvert^2-\frac{1}{N_{C}}\sum\limits_{p=1}^{K-1}\sum\limits_{q=p+1}^{K}\Arrowvert \widetilde{\mathbf{\textbf{C}}}_{p}-\widetilde{\mathbf{\textbf{C}}}_{q}\Arrowvert^2.
\end{aligned}
\end{equation}
So the objective function has another form:
\begin{equation}
\begin{aligned}
   F(\theta,\texttt{V}_{data})=&-\eta\left[\texttt{KL}(p_{0}||p_{\infty})-\texttt{KL}(p_{1}||p_{\infty})\right]\\
&+(1-\eta)\left[L_{data}(\theta)+L_{recon}(\theta)\right].
\end{aligned}
\end{equation}
\subsection{The Inference}
The next hardest problem is how to optimize the objective function:
 \begin{equation}
\begin{aligned}
  \texttt{arg}\mathop {\min }\limits_\theta F(\theta,\texttt{V}_{data})
   \end{aligned}
\end{equation}
The approximate derivative of the $\texttt{KL}(p_{0}||p_{\infty})-\texttt{KL}(p_{1}||p_{\infty})$ can be obtained by $\textbf{\texttt{\texttt{CD}}}_{1}$ learning method, so the next problems are how to get the gradients of $L_{data}(\theta)$ and $L_{recon}(\theta)$. Firstly, we compute the gradients of $L_{data}$ as follows. Because $L_{data}(\theta)$ has another equivalent form:
\begin{equation}
\begin{aligned}
   L_{data}(\theta)&=\frac{1}{N_{h}}\sum\limits_{k=1}^{K}\sum\limits_{\mathbf{\textbf{h}}_{s},\mathbf{\textbf{h}}_{t}\in\texttt{H}_{k}}(\mathbf{\textbf{h}}_{s}-\mathbf{\textbf{h}}_{t})(\mathbf{\textbf{h}}_{s}-\mathbf{\textbf{h}}_{t})^{T}\\
   &-\frac{1}{N_{C}}\sum\limits_{p=1}^{K-1}\sum\limits_{q=p+1}^{K}(\mathbf{\textbf{C}}_{p}-\mathbf{\textbf{C}}_{q})(\mathbf{\textbf{C}}_{p}-\mathbf{\textbf{C}}_{q})^T.
\end{aligned}
\end{equation}
Then we can obtain:
\begin{equation}
\begin{aligned}
  \frac {\partial L_{data}(\theta)}{\partial w_{ij}}&=\frac{1}{N_{h}}\bigg[ \sum\limits_{k=1}^{K}\sum\limits_{\mathbf{\textbf{h}}_{s},\mathbf{\textbf{h}}_{t}\in\texttt{H}_{k}}(\mathbf{\textbf{h}}_{s}-\mathbf{\textbf{h}}_{t})\frac{\partial (\mathbf{\textbf{h}}_{s}-\mathbf{\textbf{h}}_{t})^T}{\partial w_{ij}}+\\
  &\sum\limits_{k=1}^{K}\sum\limits_{\mathbf{\textbf{h}}_{s},\mathbf{\textbf{h}}_{t}\in\texttt{H}_{k}}\frac{\partial(\mathbf{\textbf{h}}_{s}-\mathbf{\textbf{h}}_{t})}{\partial w_{ij}}(\mathbf{\textbf{h}}_{s}-\mathbf{\textbf{h}}_{t})^T \bigg]-\\
  &\frac{1}{N_{C}}\bigg[\sum\limits_{p=1}^{K-1}\sum\limits_{q=p+1}^{K}
  (\mathbf{\textbf{C}}_{p}-\mathbf{\textbf{C}}_{q})\frac{\partial (\mathbf{\textbf{C}}_{p}-\mathbf{\textbf{C}}_{q})^T}{\partial w_{ij}}+\\
  &\sum\limits_{p=1}^{K-1}\sum\limits_{q=p+1}^{K}\frac{\partial(\mathbf{\textbf{C}}_{p}-\mathbf{\textbf{C}}_{q})}{\partial w_{ij}}(\mathbf{\textbf{C}}_{p}-\mathbf{\textbf{C}}_{q})^T\bigg].
\end{aligned}
\end{equation}
\indent From above result, we can see that the following task is how to compute $\frac{\partial (\mathbf{\textbf{h}}_{s}-\mathbf{\textbf{h}}_{t})^T}{\partial w_{ij}}$, $\frac{\partial(\mathbf{\textbf{h}}_{s}-\mathbf{\textbf{h}}_{t})}{\partial w_{ij}}$, $\frac{\partial (\mathbf{\textbf{C}}_{p}-\mathbf{\textbf{C}}_{q})^T}{\partial w_{ij}}$ and $\frac{\partial(\mathbf{\textbf{C}}_{p}-\mathbf{\textbf{C}}_{q})}{\partial w_{ij}}$. Next, all of them are solved separately.
\begin{equation}
\begin{aligned}
  \frac{\partial (\mathbf{\textbf{h}}_{s}-\mathbf{\textbf{h}}_{t})^T}{\partial w_{ij}}&=(\frac{\partial \mathbf{\textbf{h}}_{s}}{\partial w_{ij}}-\frac{\partial \mathbf{\textbf{h}}_{t}}{\partial w_{ij}})^T\\
  &=\bigg[\frac{\partial(\sigma(\mathbf{b}+\mathbf{\textbf{v}}_{s}\mathbf{W}))}{\partial w_{ij}}-\frac{\partial (\sigma(\mathbf{b}+\mathbf{\textbf{v}}_{t}\mathbf{W}))}{\partial w_{ij}}\bigg]^T.
\end{aligned}
\end{equation}
\indent Obviously, $\sigma(\mathbf{b}+\mathbf{\textbf{v}}_{s}\mathbf{W})$ is a row vector, all components of which are independent of $w_{ij}$ except $j$ component. So,
\begin{equation}
\begin{aligned}
  &\frac{\partial \sigma(\mathbf{b}+\mathbf{\textbf{v}}_{s}\mathbf{W})}{\partial w_{ij}}=\\
  &(\underbrace{0,\cdots,0}_{j-1},\frac{\partial \sigma(b_{j}+\sum\limits_{i=1}^{n}v_{si}w_{ij})}{\partial w_{ij}},\underbrace{0,\cdots,0}_{n-j}).\\
\end{aligned}
\end{equation}
Because $\frac{\partial \sigma(b_{j}+\sum\limits_{i=1}^{n}v_{si}w_{ij})}{\partial w_{ij}}=h_{sj}(1-h_{sj})v_{si}$, the final result of $\frac{\partial \sigma(\mathbf{b}+\mathbf{\textbf{v}}_{s}\mathbf{W})}{\partial w_{ij}}$ has an expression as follows:
\begin{equation}
\begin{aligned}
  &\frac{\partial \sigma(\mathbf{b}+\mathbf{\textbf{v}}_{s}\mathbf{W})}{\partial w_{ij}}=(\underbrace{0,\cdots,0}_{j-1},h_{sj}(1-h_{sj})v_{si},\underbrace{0,\cdots,0}_{n-j}).
\end{aligned}
\end{equation}
Similarly, the expression of final result of $\frac{\partial \sigma(\mathbf{b}+\mathbf{\textbf{v}}_{t}\mathbf{W})}{\partial w_{ij}}$ is as follows:
\begin{equation}
\begin{aligned}
  &\frac{\partial \sigma(\mathbf{b}+\mathbf{\textbf{v}}_{t}\mathbf{W})}{\partial w_{ij}}=(\underbrace{0,\cdots,0}_{j-1},h_{tj}(1-h_{tj})v_{ti},\underbrace{0,\cdots,0}_{n-j}).
\end{aligned}
\end{equation}
Then, the final result of $\frac{\partial (\mathbf{\textbf{h}}_{s}-\mathbf{\textbf{h}}_{t})^T}{\partial w_{ij}}$ is a column vector:
\begin{equation}
\begin{aligned}
  &\frac{\partial (\mathbf{\textbf{h}}_{s}-\mathbf{\textbf{h}}_{t})^T}{\partial w_{ij}}=\\
  &(\underbrace{0,\cdots,0}_{j-1},h_{sj}(1-h_{sj})v_{si}-h_{tj}(1-h_{tj})v_{ti},\underbrace{0,\cdots,0}_{n-j})^T.
\end{aligned}
\end{equation}
Similarly,
\begin{equation}
\begin{aligned}
  &\frac{\partial (\mathbf{\textbf{h}}_{s}-\mathbf{\textbf{h}}_{t})}{\partial w_{ij}}=\\
  &(\underbrace{0,\cdots,0}_{j-1},h_{sj}(1-h_{sj})v_{si}-h_{tj}(1-h_{tj})v_{ti},\underbrace{0,\cdots,0}_{n-j}),
\end{aligned}
\end{equation}
\begin{equation}
\begin{aligned}
  &\frac{\partial (\mathbf{\textbf{C}}_{p}-\mathbf{\textbf{C}}_{q})^T}{\partial w_{ij}}=\\
  &(\underbrace{0,\cdots,0}_{j-1},C_{pj}(1-C_{pj})O_{pi}-C_{qj}(1-C_{qj})O_{qi},\underbrace{0,\cdots,0}_{n-j})^T
\end{aligned}
\end{equation}
and
\begin{equation}
\begin{aligned}
  &\frac{\partial (\mathbf{\textbf{C}}_{p}-\mathbf{\textbf{C}}_{q})}{\partial w_{ij}}=\\
  &(\underbrace{0,\cdots,0}_{j-1},C_{pj}(1-C_{pj})O_{pi}-C_{qj}(1-C_{qj})O_{qi},\underbrace{0,\cdots,0}_{n-j}).
\end{aligned}
\end{equation}
Eqs. (26), (27), (28) and (29) are substituted in Eq. (21). Then
\begin{equation}
\begin{aligned}
  &\frac {\partial L_{data}(\theta)}{\partial w_{ij}}=\\
  &\frac{2}{N_{h}}\sum\limits_{k=1}^{K}\sum\limits_{\mathbf{\textbf{h}}_{s},\mathbf{\textbf{h}}_{t}\in\texttt{H}_{k}}(h_{sj}-h_{tj})\bigg[h_{sj}(1-h_{sj})v_{si}-h_{tj}(1-h_{tj})v_{ti} \bigg]\\
  &-\frac{2}{N_{C}}\sum\limits_{p=1}^{K-1}\sum\limits_{q=p+1}^{K}(C_{pj}-C_{qj})\bigg[C_{pj}(1-C_{pj})O_{pi}-\\
  &\qquad \qquad \qquad \qquad \qquad \qquad \qquad C_{qj}(1-C_{qj})O_{qi} \bigg].
\end{aligned}
\end{equation}
\indent Using above same solution, we can obtain:
\begin{equation}
\begin{aligned}
  &\frac {\partial L_{recon}(\theta)}{\partial w_{ij}}=\\
  &\frac{2}{N_{h}}\sum\limits_{k=1}^{K}\sum\limits_{\widetilde{\mathbf{\textbf{h}}}_{s},\widetilde{\mathbf{\textbf{h}}}_{t}\in\widetilde{\texttt{H}_{k}}}(\widetilde{h}_{sj}-\widetilde{h}_{tj})\bigg[\widetilde{h}_{sj}(1-\widetilde{h}_{sj})\widetilde{v}_{si}-\widetilde{h}_{tj}(1-\widetilde{h}_{tj})\widetilde{v}_{ti} \bigg]\\
  &-\frac{2}{N_{C}}\sum\limits_{p=1}^{K-1}\sum\limits_{q=p+1}^{K}(\widetilde{C}_{pj}-\widetilde{C}_{qj})\bigg[\widetilde{C}_{pj}(1-\widetilde{C}_{pj})\widetilde{o}_{pi}-\\
  &\qquad \qquad \qquad \qquad \qquad \qquad \qquad \widetilde{C}_{qj}(1-\widetilde{C}_{qj})\widetilde{o}_{qi} \bigg].
\end{aligned}
\end{equation}
\indent The following task is how to obtain $\frac {\partial L_{recon}(\theta)}{\partial b_{j}}$ and $\frac {\partial L_{recon}(\theta)}{\partial b_{j}}$. Because $\frac{\partial \sigma(b_{j}+\sum\limits_{i=1}^{n}v_{si}w_{ij})}{\partial b_{j}}=h_{sj}(1-h_{sj})$ and $\frac{\partial \sigma(b_{j}+\sum\limits_{i=1}^{n}v_{ti}w_{ij})}{\partial b_{j}}=h_{tj}(1-h_{tj})$, the final result of $\frac{\partial \sigma(\mathbf{b}+\mathbf{\textbf{v}}_{s}\mathbf{W})}{\partial b_{j}}$ and $\frac{\partial \sigma(\mathbf{b}+\mathbf{\textbf{v}}_{t}\mathbf{W})}{\partial b_{j}}$ have expressions as follows:
\begin{equation}
\begin{aligned}
  &\frac{\partial \sigma(\mathbf{b}+\mathbf{\textbf{v}}_{s}\mathbf{W})}{\partial b_{j}}=(\underbrace{0,\cdots,0}_{j-1},h_{sj}(1-h_{sj}),\underbrace{0,\cdots,0}_{n-j})
\end{aligned}
\end{equation}
and
\begin{equation}
\begin{aligned}
  &\frac{\partial \sigma(\mathbf{b}+\mathbf{\textbf{v}}_{t}\mathbf{W})}{\partial b_{j}}=(\underbrace{0,\cdots,0}_{j-1},h_{tj}(1-h_{tj}),\underbrace{0,\cdots,0}_{n-j}).
\end{aligned}
\end{equation}
So, the final result of $\frac {\partial L_{data}(\theta)}{\partial b_{j}}$ is as follows.
\begin{equation}
\begin{aligned}
  &\frac {\partial L_{data}(\theta)}{\partial b_{j}}=\\
  &\frac{2}{N_{h}}\sum\limits_{k=1}^{K}\sum\limits_{\mathbf{\textbf{h}}_{s},\mathbf{\textbf{h}}_{t}\in\texttt{H}_{k}}(h_{sj}-h_{tj})\bigg[h_{sj}(1-h_{sj})-h_{tj}(1-h_{tj}) \bigg]\\
  &-\frac{2}{N_{C}}\sum\limits_{p=1}^{K-1}\sum\limits_{q=p+1}^{K}(C_{pj}-C_{qj})\bigg[C_{pj}(1-C_{pj})-\\
  &\qquad \qquad \qquad \qquad \qquad \qquad \qquad C_{qj}(1-C_{qj}) \bigg].
\end{aligned}
\end{equation}
Similarly, the expression of $\frac {\partial L_{recon}(\theta)}{\partial b_{j}}$ is as follows.
\begin{equation}
\begin{aligned}
  &\frac {\partial L_{recon}(\theta)}{\partial b_{j}}=\\
  &\frac{2}{N_{h}}\sum\limits_{k=1}^{K}\sum\limits_{\widetilde{\mathbf{\textbf{h}}}_{s},\widetilde{\mathbf{\textbf{h}}}_{t}\in\widetilde{\texttt{H}}_{k}}(\widetilde{h}_{sj}-\widetilde{h}_{tj})\bigg[\widetilde{h}_{sj}(1-\widetilde{h}_{sj})-\widetilde{h}_{tj}(1-\widetilde{h}_{tj}) \bigg]\\
  &-\frac{2}{N_{C}}\sum\limits_{p=1}^{K-1}\sum\limits_{q=p+1}^{K}(\widetilde{C}_{pj}-\widetilde{C}_{qj})\bigg[\widetilde{C}_{pj}(1-\widetilde{C}_{pj})-\\
  &\qquad \qquad \qquad \qquad \qquad \qquad \qquad \widetilde{C}_{qj}(1-\widetilde{C}_{qj}) \bigg].
\end{aligned}
\end{equation}
\indent Because $\frac{\partial \sigma(b_{j}+\sum\limits_{i=1}^{n}v_{si}w_{ij})}{\partial a_{i}}=0$ and $\frac{\partial \sigma(b_{j}+\sum\limits_{i=1}^{n}v_{ti}w_{ij})}{\partial a_{i}}=0$, the final results of $\frac{\partial \sigma(\mathbf{b}+\mathbf{\textbf{v}}_{s}\mathbf{W})}{\partial a_{i}}$ and $\frac{\partial \sigma(\mathbf{b}+\mathbf{\textbf{v}}_{t}\mathbf{W})}{\partial a_{i}}$ are zero vectors. Then we can obtain: $ \frac {\partial L_{data}(\theta)}{\partial a_{i}}=0, \frac {\partial L_{recon}(\theta)}{\partial a_{i}}=0$.\\
\indent Finally, following the $\textbf{\texttt{CD}}_{1}$ learning and the gradient of the $L_{data}$ and $L_{recon}$, the update rules of the symmetric connection weight take the form:
\begin{equation}
\begin{aligned}
   &w_{ij}^{(\tau+1)}=w_{ij}^{(\tau)}+\eta\varepsilon(<v_{i}h_{j}>_{0}-<v_{i}h_{j}>_{1})+(1-\eta)\\ &\Bigg\{\frac{2}{N_{h}}\sum\limits_{k=1}^{K}\sum\limits_{\mathbf{\textbf{h}}_{s},\mathbf{\textbf{h}}_{t}\in\texttt{H}_{k}}(h_{sj}-h_{tj})\bigg[h_{sj}(1-h_{sj})v_{si}-h_{tj}(1-\\
   &h_{tj})v_{ti} \bigg]-\frac{2}{N_{C}}\sum\limits_{p=1}^{K-1}\sum\limits_{q=p+1}^{K}(C_{pj}-C_{qj})\bigg[C_{pj}(1-C_{pj})O_{pi}\\
   &-C_{qj}(1-C_{qj})O_{qi} \bigg]\Bigg\}+(1-\eta)\Bigg\{\frac{2}{N_{h}}\sum\limits_{k=1}^{K}\sum\limits_{\widetilde{\mathbf{\textbf{h}}}_{s},\widetilde{\mathbf{\textbf{h}}}_{t}\in\widetilde{\texttt{H}_{k}}}(\widetilde{h}_{sj}\\
   &-\widetilde{h}_{tj})\bigg[\widetilde{h}_{sj}(1-\widetilde{h}_{sj})\widetilde{v}_{si}-\widetilde{h}_{tj}(1-\widetilde{h}_{tj})\widetilde{v}_{ti} \bigg]-\frac{2}{N_{C}}\sum\limits_{p=1}^{K-1}\sum\limits_{q=p+1}^{K}\\
   &(\widetilde{C}_{pj}-\widetilde{C}_{qj})\bigg[\widetilde{C}_{pj}(1-\widetilde{C}_{pj})\widetilde{o}_{pi}-\widetilde{C}_{qj}(1-\widetilde{C}_{qj})\widetilde{o}_{qi} \bigg]\Bigg\}.
   \end{aligned}
\end{equation}
The update rules of biases $b_{j}$ take the form:
\begin{equation}
\begin{aligned}
   &b_{j}^{(\tau+1)}=b_{j}^{(\tau)}+\eta\varepsilon(<h_{j}>_{0}-<h_{j}>_{1})+(1-\eta)\\ &\Bigg\{\frac{2}{N_{h}}\sum\limits_{k=1}^{K}\sum\limits_{\mathbf{\textbf{h}}_{s},\mathbf{\textbf{h}}_{t}\in\texttt{H}_{k}}(h_{sj}-h_{tj})\bigg[h_{sj}(1-h_{sj})-h_{tj}(1-h_{tj}) \bigg]\\
  &-\frac{2}{N_{C}}\sum\limits_{p=1}^{K-1}\sum\limits_{q=p+1}^{K}(C_{pj}-C_{qj})\bigg[C_{pj}(1-C_{pj})-\\
  &\qquad \qquad \qquad \qquad \qquad \qquad \qquad C_{qj}(1-C_{qj}) \bigg]\Bigg\}+(1-\eta)\\
  &\Bigg\{\frac{2}{N_{h}}\sum\limits_{k=1}^{K}\sum\limits_{\widetilde{\mathbf{\textbf{h}}}_{s},\widetilde{\mathbf{\textbf{h}}}_{t}\in\widetilde{\texttt{H}_{k}}}(\widetilde{h}_{sj}-\widetilde{h}_{tj})\bigg[\widetilde{h}_{sj}(1-\widetilde{h}_{sj})-\widetilde{h}_{tj}(1-\widetilde{h}_{tj}) \bigg]\\
  &-\frac{2}{N_{C}}\sum\limits_{p=1}^{K-1}\sum\limits_{q=p+1}^{K}(\widetilde{C}_{pj}-\widetilde{C}_{qj})\bigg[\widetilde{C}_{pj}(1-\widetilde{C}_{pj})-\\
  &\qquad \qquad \qquad \qquad \qquad \qquad \qquad \widetilde{C}_{qj}(1-\widetilde{C}_{qj}) \bigg]\Bigg\}
   \end{aligned}
\end{equation}
and the update rules of biases $a_{i}$ take the form:
\begin{equation}
\begin{aligned}
a_{i}^{(\tau+1)}=a_{i}^{(\tau)}+\eta\varepsilon(<v_{i}>_{0}-<v_{i}>_{1}).
   \end{aligned}
\end{equation}
\indent Next, for iterative updating rules Eqs. (36), (37) and (38), the convergence needs to be analyzed theoretically.
\subsection{Convergence Analysis}
In this subsection, we use the convergence analysis method \cite{Yang2018Feature}, \cite{7872408} to prove that the objective function decreases monotonically by the iterative updating rules Eqs. (36), (37) and (38).\\
\indent \textbf{Theorem 1:} In each iteration, the objective function Eq. (15) will monotonically decrease. \\
\indent \textbf{Proof:} Assume that $\textbf{a}^{\tau}$, $\textbf{b}^{\tau}$ and $\textbf{W}^{\tau}$ have been derived in the $\tau$th iteration. We fix $\textbf{a}^{\tau}$ and $\textbf{b}^{\tau}$ to optimize $\textbf{W}^{\tau}$. Obviously, according to stochastic gradient descent method, there is the following inequality:
\begin{equation}
\begin{aligned}
F(\textbf{a}^{\tau},\textbf{b}^{\tau},\textbf{W}^{(\tau+1)},\texttt{V}_{data})\leq F(\textbf{a}^{\tau},\textbf{b}^{\tau},\textbf{W}^{\tau},\texttt{V}_{data})
   \end{aligned}
\end{equation}
Then, the parameters $\textbf{W}$ and $\textbf{a}$ are fixed as $\textbf{W}^{(\tau+1)}$ and $\textbf{a}^{\tau}$ to optimize $\textbf{b}$. We have another inequality:
\begin{equation}
\begin{aligned}
F(\textbf{a}^{\tau},\textbf{b}^{(\tau+1)}, \textbf{W}^{(\tau+1)},\texttt{V}_{data})\leq F(\textbf{a}^{\tau},\textbf{b}^{\tau},\textbf{W}^{(\tau+1)},\texttt{V}_{data})
   \end{aligned}
\end{equation}
Similarily, the parameters $\textbf{W}$ and $\textbf{b}$ are fixed as $\textbf{W}^{(\tau+1)}$ and $\textbf{b}^{(\tau+1)}$ to optimize $\textbf{a}$. We get the third inequality:
\begin{equation}
\begin{aligned}
F(\textbf{a}^{(\tau+1)},\textbf{b}^{(\tau+1)}, \textbf{W}^{(\tau+1)},\texttt{V}_{data})\leq F(\textbf{a}^{\tau},\textbf{b}^{\tau+1},\textbf{W}^{\tau+1},\texttt{V}_{data})
   \end{aligned}
\end{equation}
So, from Eqs. (39), (40) and (41), we have
\begin{equation}
\begin{aligned}
F(\textbf{a}^{\tau+1},\textbf{b}^{\tau+1}, \textbf{W}^{\tau+1},\texttt{V}_{data})\leq F(\textbf{a}^{\tau},\textbf{b}^{\tau}, \textbf{W}^{\tau},\texttt{V}_{data})
   \end{aligned}
\end{equation}
Thus, the objective function decreases monotonically by the iterative updating rules Eqs. (36), (37) and (38). The learning algorithm of the MIRBM model is summarized as follows.\\\\
\textbf{Learning Algorithm of MIRBM}\\
\noindent\line(1,0){250}\\
\textbf{Input}: $\texttt{V}_{data}$ is original data set;\\
\indent \indent\indent $\eta$ is a scale coefficient;\\
\indent \indent\indent $\texttt{LCP}$ is the local cluster partition of the $\texttt{V}_{data}$.\\
\noindent\line(1,0){250}\\
\textbf{Output}: The model parameters of $\textbf{a}$,$\textbf{b}$ and $\textbf{W}$.\\
\noindent\line(1,0){250}\\
\indent \textbf{For} each iteration \textbf{do}\\
\indent \indent \textbf{For} all visible layer vectors \textbf{do}\\
\indent \indent \indent Sample hidden layer distribution by Eq. (3);\\
\indent \indent \textbf{End For} \\
\indent \indent \textbf{For} all hidden layer vectors \textbf{do}\\
\indent \indent \indent \textbf{If} visible layer is binary unit \textbf{do}\\
\indent \indent \indent \indent Sample reconstructed visible layer distribution by\\
 \indent \indent \indent \indent Eq. (4);\\
\indent \indent \indent \textbf{Else}\\
\indent \indent \indent \indent Sample reconstructed visible layer distribution by \\
\indent \indent \indent \indent linear transformation;\\
\indent \indent \textbf{End For} \\
\indent \indent Obtain the gradients of Eqs. (30), (31), (34) and (35);\\
\indent \indent Use Eqs. (36), (37) and (38) to update $\textbf{W}$, $\textbf{b}$ and \\
\indent \indent  $\textbf{a}$, respectively;\\
\indent \textbf{End For}\\
\indent  \textbf{Return} $\mathbf{W}$, $\mathbf{b}$ and $\mathbf{a}$.\\
\noindent\line(1,0){250}
\subsection{Complexity Analysis}
Let the iteration times of the MIRBM model learning algorithm be $T_{s}$ and $\texttt{V}_{data}$ have $N$ vectors. Then, the time complexities of hidden layer feature sampling and reconstructed visible layer distribution sampling are both $O(N)$. Suppose that the auxiliary LCP guiding has $K$ clusters and the $k(k=1,2,\cdots,K)$ cluster has $M_{k}$ instances. To obtain the gradients of Eqs. (30), (31), (34) and (35), the time complexities are both $O(K\frac{M_{k}(M_{k}-1)}{2})$. Other step of MIRBM model learning algorithm need $O(1)$ time complex. Eventually, the time complexity of MIRBM model learning algorithm is $O(T_{s}(N+KM_{k}(M_{k}-1)))$.
\section{Experiment Results}
In this section, we compare the proposed unsupervised feature learning architecture with the state-of-the-art GraphRBM using image clustering task. The experimental data include fifteen real-valued image datasets. Three groups nine algorithms are used to prove the hidden feature learning capability and generalization ability of our architecture. We choose four external evaluation metrics to evaluate the clustering performance.
\subsection{Datasets}
\indent To compare with the state-of-the-art GraphRBM, we choose fifteen real-valued image datasets from  from the Microsoft Research Asia Multimedia (MSRA-MM)2.0 \cite{li2009msra}. The summary of the experiment datasets is shown in Table II.
\begin{table}
\begin{center}
\caption{Summary of the experiment datasets.}
 \label{imagedata}
\scalebox{1}{
\begin{tabular}{lcccc}
\toprule[1.5pt] 
{No.} &\textsf{\bf{Dataset}} &  {classes}&  {Instances} &  {feature} \\
\hline
{1} & \textsf{alphabet} & {3} & {814}  & {892} \\
{2} &\textsf{americanflag} & {3} & {873} &{892} \\
{3} & \textsf{aquarium} & {3} & {922} &{892} \\
{4} &\textsf{baobab} & {3} & {900}  & {892} \\
{5} &\textsf{bathroom}& {3} & {924} &{892}\\
{6} &\textsf{beret} & {3} & {876} &{892} \\
{7} &\textsf{birthdaycake} & {3} & {932} &{892} \\
{8} &\textsf{blog} & {3} & {943} &{892} \\
{9} &\textsf{blood} & {3} & {866} &{892} \\
{10} &\textsf{boat}& {3} & {857} &{892}\\
{11} &\textsf{boomerang} & {3} & {910} &{892} \\
{12} &\textsf{building} & {3} & {911} &{892} \\
{13} &\textsf{wallpaper} & {3} & {919} &{899} \\
{14} &\textsf{weapon} & {3} & {858} &{899} \\
{15} &\textsf{weddingdress} & {3} & {883} &{899} \\
\bottomrule[1pt] 
\end{tabular}}
\end{center}
\end{table}
\subsection{Experimental Settings}
\indent To compare our architecture with the state-of-the-art GraphRBM, we choose fifteen real-valued image datasets in the experiments. In the MIRBM model of the proposed architecture, the visible layer units are set to Gaussian linear visible units for real-valued data sets. A linear transformation is adopted in the process of visible layer reconstruction. To compare fairness, the learning rate of our MIRBM model and the GraphRBM is both set to $10^{-2}$. One of the key parameters $\eta$ is set to $0.1,0.2,\cdots,0.9$ to analysis the sensitivity.\\
\indent We choose three groups nine clustering algorithms to evaluate the hidden layer feature learning ability and generalization ability of our feature learning architecture. The first group is K-means \cite{Lloyd1982Least}, affinity propagation (AP) \cite{frey2007clustering} and spectral clustering (SC)\cite{ng2002spectral} algorithms using the original data as their input. The second group is K-means+GraphRBM, AP+GraphRBM and SC+GraphRBM algorithms based on the GraphRBM using the hidden layer features of the GraphRBM as their input. The last group is K-means+MIRBM, AP+MIRBM and SC+MIRBM algorithms based on our architecture using the hidden layer features of the MIRBM model as their input.
\subsection{Evaluation Metrics}
\indent To prove the hidden layer feature learning capability of the proposed architecture, we choose four external evaluation metrics accuracy \cite{Cai2005Document}, purity \cite{ding2006orthogonal}, Fowlkes and Mallows Index \cite{LIU2018200} and Jaccard Index\cite{Real1996The} \cite{Jaccard2018index}) to evaluate the clustering performance.
\subsubsection{Accuracy}
Given an instance $x_{i}$, let $h_{i}$ and $g_{i}$ be the cluster label and the true label, respectively. The clustering accuracy is given by:
\begin{equation}
\begin{aligned}
    AC=\frac{\sum\limits_{i=1}\delta(g_{i},map(h_{i}))}{n},
\end{aligned}
\end{equation}
where $AC\in[0,1]$, $n$ is the total number of instances, $map(h_{i})$ maps each cluster label $h_{i}$ to the equivalent label from the data set, and $\delta(x,y)$ equals to one if $x=y$ and equals to zero otherwise.

\subsubsection{Purity}
\indent The purity is used to measure the extent of each cluster contained data points from primarily one class. It is an external transparent evaluation measure for cluster quality. The purity is given by:
 \begin{equation}
  purity=\sum_{i=1}^K\frac{n_i}{n}P(S_i), P(S_i)=\frac{1}{n_i}\max_j(n_i^j),
 \end{equation}
  where $n_i^j$  is the number of the $i$-th input class that is assigned to the $j$-th cluster and $S_i$ is a particular cluster size of $n_i$.\\
\subsubsection{Fowlkes and Mallows Index}
The Fowlkes and Mallows index is an external evaluation method for two clusterings that can be defined as:
\begin{equation}
\begin{aligned}
  \texttt{\textbf{FMI}}=\sqrt{\frac{TP}{TP+FP}\times\frac{TP}{TP+FN}},
   \end{aligned}
\end{equation}
where $TP$, $FP$ and $FN$ are the number of true positives, false positives and false negatives, respectively.
\subsubsection{Jaccard Index}
The Jaccard Index as an external evaluation method is used to gauge the diversity and similarity of sample sets. Given two sets $A$ and $B$, each of them has $n$ attributes. The total number of each combination of attributes is defined by:
\begin{itemize}
  \item $M_{a}$ represents the total number of attributes, if $A$ and $B$ have a value of 1.
  \item $M_{b}$ represents the total number of attributes, if the attribute of $A$ has a value 0 and $B$ has a value of 1.
  \item $M_{c}$ represents the total number of attributes, if the attribute of $A$ has a value 1 and $B$ has a value of 0.
\end{itemize}
The Jaccard Index, $J$, is given as:
\begin{equation}
\begin{aligned}
  J=\frac{M_{a}}{M_{b}+M_{c}+M_{a}}.
   \end{aligned}
\end{equation}
\subsection{Performance Comparison}
In this subsection, we use three groups of nine clustering algorithms to compare the state-of-the-art GraphRBM and our unsupervised feature learning architecture.
\subsubsection{Accuracy Comparison}
In Table III, we list the results of the accuracy of three groups contrastive algorithms. The first group is K-means, AP and SC algorithm using original data as input. The second group is K-means+GraphRBM, AP+GraphRBM and SC+GraphRBM algorithm using the hidden layer features of the GraphRBM as input. The final group is K-means+MIRBM, AP+MIRBM and SC+MIRBM algorithm using the hidden layer features of our MIRBM model as input. From Table III, we can see that the best performances distribute over the algorithms based on our architecture. For some datasets (e.g. blog, boomerang, wallpaper and weddingdress), K-means+MIRBM algorithm shows the best performance. And the SC+MIRBM algorithm shows the best performance in the datasets baobab and birthdaycake. For other datasets, the AP+MIRBM algorithm shows the best performance. For all datasets, the average accuracies of K-means, AP and SC algorithms are 0.4309, 0.4408 and 0.4194, respectively. The K-means+GraphRBM, AP+GraphRBM and SC+GraphRBM algorithms based on the GraphRBM raise the average accuracies to 0.4584, 0.4623 and 0.4472, respectively. However, the average accuracies of K-means+MIRBM, AP+MIRBM and SC+MIRBM algorithms based on our architecture raise greatly to 0.5783, 0.6031 and 0.5273, respectively. Hence, the hidden layer features of MIRBM model as the input of K-means, AP and SC algorithms can enhance the average accuracies by 14.71\%, 16.23\% and 10.79\%, respectively. In terms of feature representation ability of our architecture and the GraphRBM, the average accuracies of the K-means+MIRBM, AP+MIRBM, SC+MIRBM algorithms are 11.99\%, 14.08\% and 8.01\% higher than K-means+GraphRBM, AP+GraphRBM, SC+GraphRBM algorithms, respectively.\\
\indent For each dataset, more intuitive performance comparisons are shown in Fig. 3. There are twelve datasets (alphabet, americanflag, aquarium, baobab, bathroom, beret, birthdaycake, blog, boat, boomerang, building and wallpaper) show outstanding performances about three kinds of algorithms based on our architecture simultaneously.
\subsubsection{Purity Comparison}
We report the results of the purity of K-means+GraphRBM, AP+GraphRBM, SC+GraphRBM algorithms based on the GraphRBM and K-means+MIRBM, AP+MIRBM, SC+MIRBM algorithms based on our architecture in Table IV. From Table IV, we can see that the best performances distribute mainly over the algorithms based on our architecture expect for the baobab and birthdaycake datasets. The results of average purity of K-means, AP and SC algorithms using original data as input are 0.7483, 0.7509 and 0.7465, respectively. But, the results of average purity of K-means+GraphRBM, AP+GraphRBM and SC+GraphRBM algorithms using the hidden features of GraphRBM as input are raised to 0.7578, 0.7625 and 0.7568, respectively. For our architecture, the results of average Jac of K-means+MIRBM, AP+MIRBM and SC+MIRBM algorithms using the hidden features of it as input are significantly raised to 0.7590, 0.7653 and 0.7607, respectively. As a whole, the proposed architecture has better feature learning ability than the GraphRBM.\\
\indent Among three groups contrastive algorithms, the intuitive comparisons of the average purity results are shown in Fig. 4. All the algorithms based on our MIRBM model show better performances.
\subsubsection{FMI Comparison}
The FMI results of all contrastive algorithms are listed in Table V. From Table V, the results of average FMI of K-means, AP and SC algorithms using original data as input are 0.4382, 0.4359 and 0.4238, respectively. They are raised to 0.4619, 0.4653 and 0.4529, respectively, by the GraphRBM using hidden layer features of it as input. However, our architecture further improves the results of average FMI to 0.5977, 0.6322 and 0.5482, respectively. Clearly, the GraphRBM and our architecture both have powerful feature learning capabilities. But, our architecture model shows more powerful hidden layer feature extraction capabilities than the GraphRBM. From Table V, the best performances of all datasets are distributed over the algorithms based on the proposed architecture, especially with regard to AP+MIRBM algorithm.\\
\indent The intuitive comparisons of the average FMI results among three groups contrastive algorithms are shown in Fig. 4. The K-means+MIRBM, AP+MIRBM and SC+MIRBM algorithms based on the proposed architecture show absolute advantages.
\subsubsection{Jac Comparison}
In Table VI, we report the results of the Jac of K-means+GraphRBM, AP+GraphRBM, SC+GraphRBM algorithms based on the GraphRBM and K-means+MIRBM, AP+MIRBM, SC+MIRBM algorithms based on the proposed architecture. Meanwhile, we report the results of the Jac of original K-means, AP and SC algorithms. From Table VI, we can see that the best performances of all datasets distribute over the algorithms based on our architecture. In fact, they distribute mainly over the algorithms AP+MIRBM. For all datasets, the results of average Jac of K-means, AP and SC algorithms using original data as input are 0.2748, 0.2749 and 0.2618, respectively. But, the results of average Jac of K-means+GraphRBM, AP+GraphRBM and SC+GraphRBM algorithms using the hidden features of GraphRBM as input are raised to 0.2989, 0.3018 and 0.2905, respectively. Hence, the GraphRBM has certain capability of feature representation. For the proposed architecture, the results of average Jac of K-means+MIRBM, AP+MIRBM and SC+MIRBM algorithms using the hidden features of it as input are significantly raised to 0.4098, 0.4463 and 0.3702, respectively. Therefore, the proposed architecture shows excellent performance for Jac evaluation metric.\\
\indent The intuitive comparisons of the average Jac results among three groups contrastive algorithms are shown in Fig. 4. All the algorithms based on the proposed architecture show absolute advantages.
\begin{table*}
\begin{center}
\caption{The performances of the accuracies and variance ($\eta=0.1$).}
\label{tab:results1} \scalebox{0.76}{
\begin{tabular}{|l|c|c|c|c|c|c|c|c|c|}
\hline
\textsf{ \bf{Dataset (No.)}} &  {K-means} & {AP} &{SC} &{K-means+GraphRBM}& {AP+GraphRBM} & {SC+GraphRBM}&{K-means+MIRBM}& {AP+MIRBM} & {SC+MIRBM}\\
\hline \textsf{alphabet} & {0.4066$\pm 0.0002 $} & {	0.4042$\pm 0.0005 $} & {	0.4386$\pm 0.0001 $} & {	0.3808$\pm 0.0003 $} & {	 0.3796$\pm 0.0001 $} & {	0.3980$\pm 0.0002 $} & {	0.4530$\pm 0.0003 $} & {	\textbf{0.4607$\pm 0.0002 $}} & {	0.4447$\pm 0.0001 $}  \\
\hline \textsf{americanflag} & {0.4513$\pm 0.0003 $} & {	0.4376$\pm 0.0001 $} & {	0.4204$\pm 0.0002 $} & {	0.4284$\pm 0.0001 $} & {	 0.4318$\pm 0.0002 $} & {	0.4124$\pm 0.0003 $} & {	0.5470$\pm 0.0004 $} & {	\textbf{	0.5670$\pm 0.0001 $}} & {	0.5384$\pm 0.0005 $}  \\
\hline \textsf{aquarium} & {0.3720$\pm 0.0001 $} & {	0.3709$\pm 0.0002 $} & {	0.3970$\pm 0.0002 $} & {	0.5293$\pm 0.0001 $} & {	 0.5315$\pm 0.0003 $} & {	0.5119$\pm 0.0001 $} & {	0.6817$\pm 0.0001 $} & {		\textbf{0.7017$\pm 0.0003 $}} & {	0.6941$\pm 0.0002 $} \\
\hline \textsf{baobab} & {0.4989$\pm 0.0002 $} & {	0.5511$\pm 0.0004 $} & {	0.4822$\pm 0.0001 $} & {	0.4722$\pm 0.0006 $} & {	 0.4778$\pm 0.0003 $} & {	 0.4456$\pm 0.0002 $} & {	0.5900$\pm 0.0004 $} & {		0.5478$\pm 0.0002 $} & {	\textbf{0.6078$\pm 0.0005 $}}\\
\hline \textsf{bathroom} & {0.3723$\pm 0.0003 $} & {	0.4221$\pm 0.0001 $} & {	0.3820$\pm 0.0004 $} & {	0.4383$\pm 0.0002 $} & {	 0.4372$\pm 0.0002 $} & {	0.4058$\pm 0.0003 $} & {	0.7787$\pm 0.0004 $} & {		\textbf{0.7987$\pm 0.0001 $}} & {	0.4286$\pm 0.0002 $}
 \\
\hline \textsf{beret} & {0.4635$\pm 0.0001 $} & {	0.3995$\pm 0.0004 $} & {	0.4110$\pm 0.0003 $} & {	0.4532$\pm 0.0002 $} & {	 0.4543$\pm 0.0003 $} & {	 0.4338$\pm 0.0003 $} & {	0.5097$\pm 0.0005 $} & {		\textbf{0.5559$\pm 0.0003 $}} & {	0.4532$\pm 0.0002 $}  \\
\hline \textsf{birthdaycake}&  {0.4957$\pm 0.0003 $} & {	0.4753$\pm 0.0001 $} & {	0.4818$\pm 0.0001 $} & {	0.5697$\pm 0.0004 $} & {	 0.5504$\pm 0.0003 $} & {	0.5322$\pm 0.0001 $} & {	0.6506$\pm 0.0002 $} & {		\textbf{0.6620$\pm 0.0002 $}} & {	\textbf{0.6620$\pm 0.0004 $}} \\
\hline \textsf{blog} & {0.3998$\pm 0.0001 $} & {	0.3860$\pm 0.0003 $} & {	0.3796$\pm 0.0003 $} & {	0.4973$\pm 0.0002 $} & {	 0.4719$\pm 0.0001 $} & {	 0.4677$\pm 0.0003 $} & {	\textbf{0.7149$\pm 0.0004 $}} & {		0.6257$\pm 0.0002 $} & {	0.5440$\pm 0.0005 $} \\
\hline \textsf{blood} & {0.4885$\pm 0.0001 $} & {	0.5589$\pm 0.0003 $} & {	0.3868$\pm 0.0004 $} & {	0.6085$\pm 0.0002 $} & {	 0.6282$\pm 0.0002 $} & {	 0.5878$\pm 0.0004 $} & {	0.4708$\pm 0.0003 $} & {		\textbf{0.8072$\pm 0.0005 $}} & {	0.4908$\pm 0.0001 $}  \\
\hline \textsf{boat} & {0.3991$\pm 0.0002 $} & {	0.4236$\pm 0.0002 $} & {	0.4014$\pm 0.0001 $} & {	0.3711$\pm 0.0003 $} & {	 0.4072$\pm 0.0001 $} & {	 0.3967$\pm 0.0002 $} & {	0.6124$\pm 0.0004 $} & {		\textbf{0.6289$\pm 0.0003 $}} & {	0.6056$\pm 0.0003 $}  \\
\hline \textsf{boomerang} & {0.3890$\pm 0.0003 $} & {	0.4923$\pm 0.0002 $} & {	0.4473$\pm 0.0004 $} & {	0.3989$\pm 0.0005 $} & {	 0.4143$\pm 0.0003 $} & {	0.3945$\pm 0.0002 $} & {	\textbf{0.5262$\pm 0.0004 $}} & {		0.5253$\pm 0.0003 $} & {	0.4681$\pm 0.0003 $}  \\
\hline \textsf{building} & {0.5148$\pm 0.0002 $} & {	0.4874$\pm 0.0001 $} & {	0.4742$\pm 0.0003 $} & {	0.4632$\pm 0.0002 $} & {	 0.4841$\pm 0.0002 $} & {	0.4479$\pm 0.0001 $} & {	0.6781$\pm 0.0003 $} & {		\textbf{0.7146$\pm 0.0001 $}} & {	0.6981$\pm 0.0004 $} \\
\hline \textsf{wallpaper} & {0.4255$\pm 0.0003 $} & {	0.4342$\pm 0.0002 $} & {	0.3874$\pm 0.0001 $} & {	0.4363$\pm 0.0002 $} & {	 0.4407$\pm 0.0002 $} & {	0.4385$\pm 0.0001 $} & {	\textbf{0.5426$\pm 0.0003 $}} & {		0.5256$\pm 0.0003 $} & {	0.4755$\pm 0.0002 $} \\
\hline \textsf{weapon} & {0.4196$\pm 0.0003 $} & {	0.3753$\pm 0.0002 $} & {	0.4254$\pm 0.0002 $} & {	0.4033$\pm 0.0004 $} & {	 0.4079$\pm 0.0006 $} & {	 0.3939$\pm 0.0002 $} & {	0.4660$\pm 0.0005 $} & {	\textbf{	0.4860$\pm 0.0006 $}} & {	0.3951$\pm 0.0004 $}  \\
\hline \textsf{weddingdress} & {0.3669$\pm 0.0002 $} & {	0.3930$\pm 0.0002 $} & {	0.3760$\pm 0.0001 $} & {	0.4258$\pm 0.0004 $} & {	 0.4179$\pm 0.0003 $} & {	0.4417$\pm 0.0002 $} & {	\textbf{0.4534$\pm 0.0002 $}} & {		0.4394$\pm 0.0004 $} & {	0.4032$\pm 0.0001 $}  \\
\hline
\hline
\hline  {Average} & {0.4309 } & {	0.4408 } & {	0.4194 } & {	0.4584 } & {	0.4623 } & {	0.4472 } & {	\textbf{0.5783} } & {		 \textbf{0.6031} } & {	\textbf{0.5273} }\\
\hline
\end{tabular}}
\end{center}
\end{table*}
\begin{figure*}
\centering
     \includegraphics[scale=0.4]{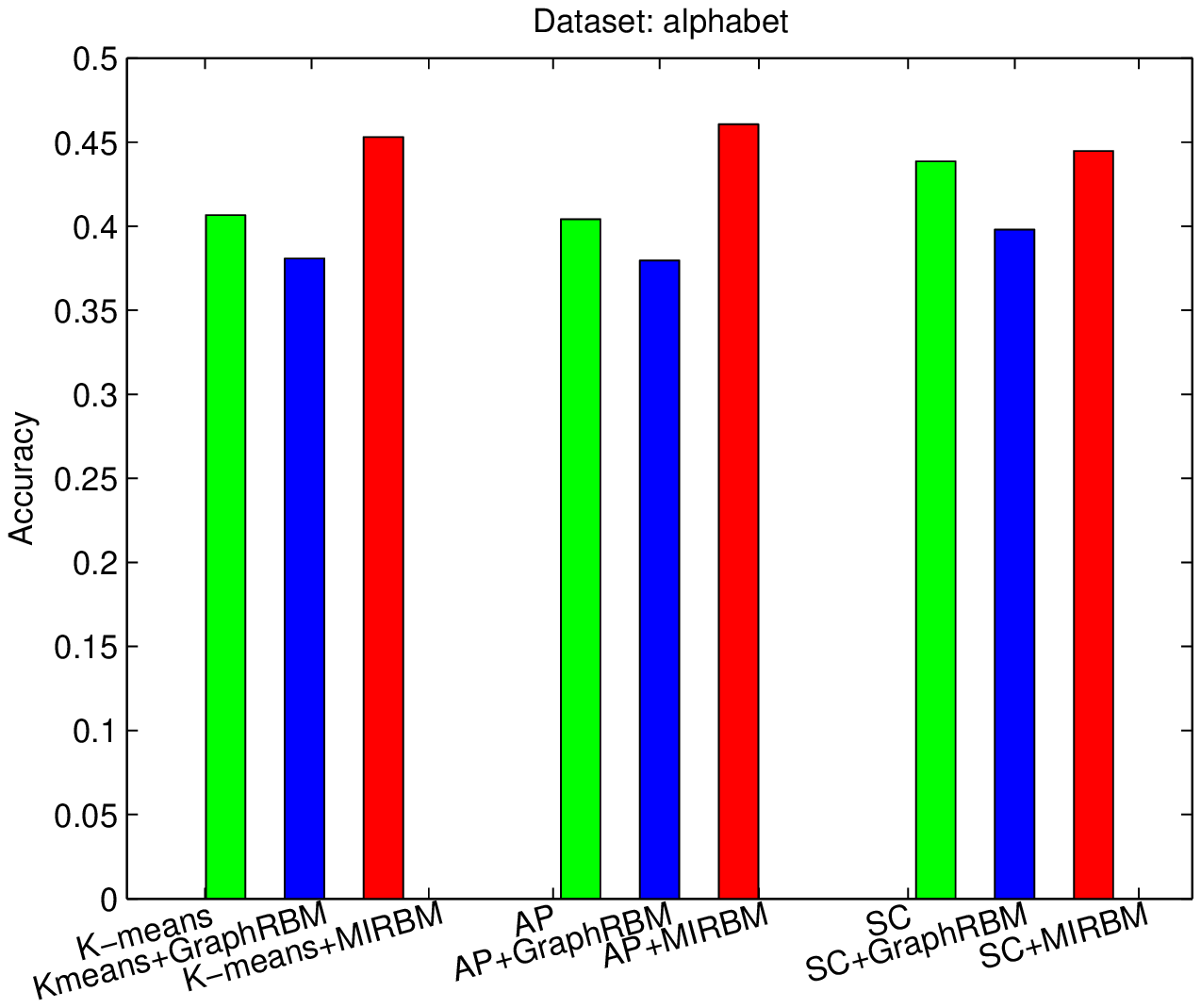}
    \includegraphics[scale=0.4]{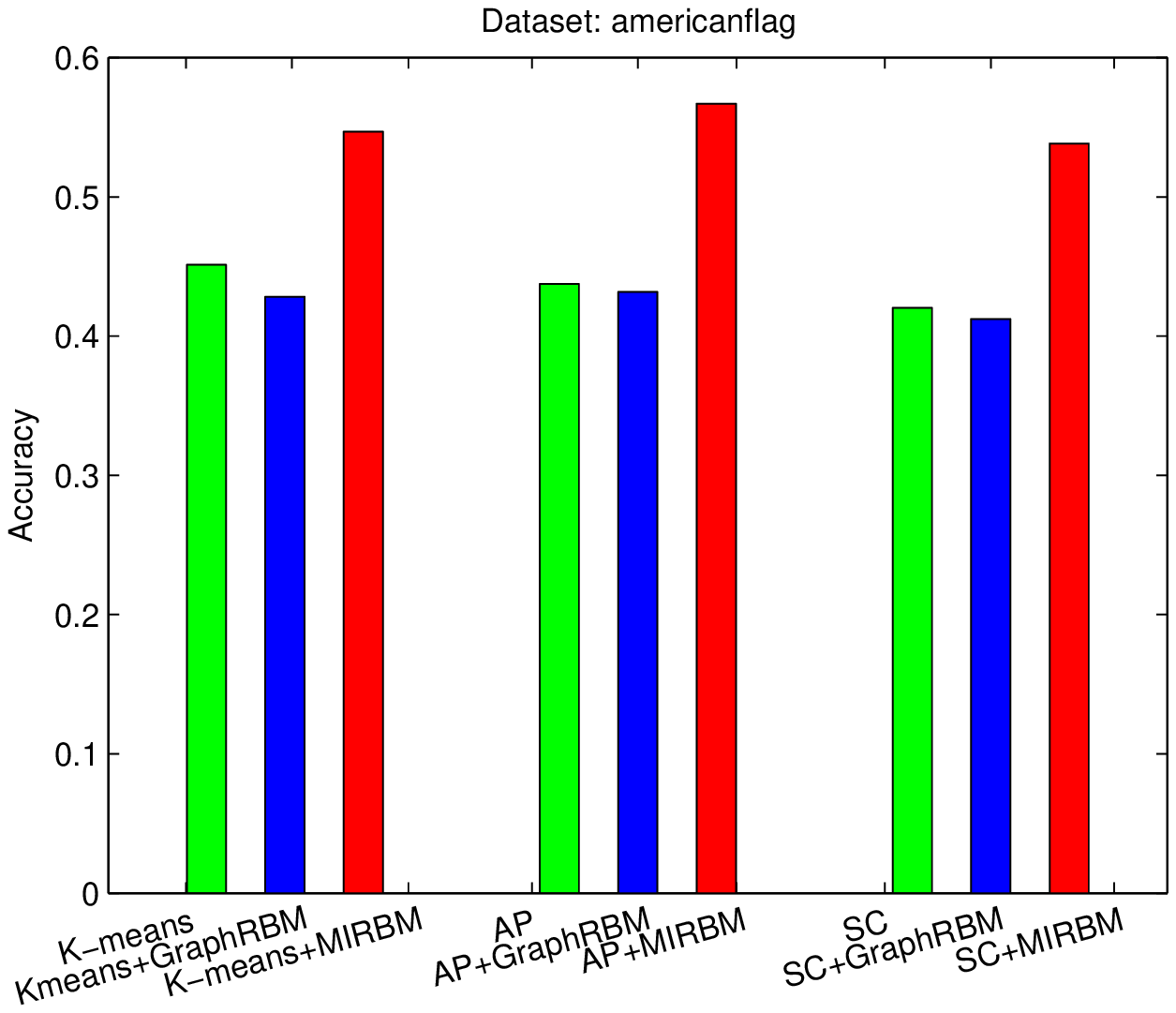}
     \includegraphics[scale=0.4]{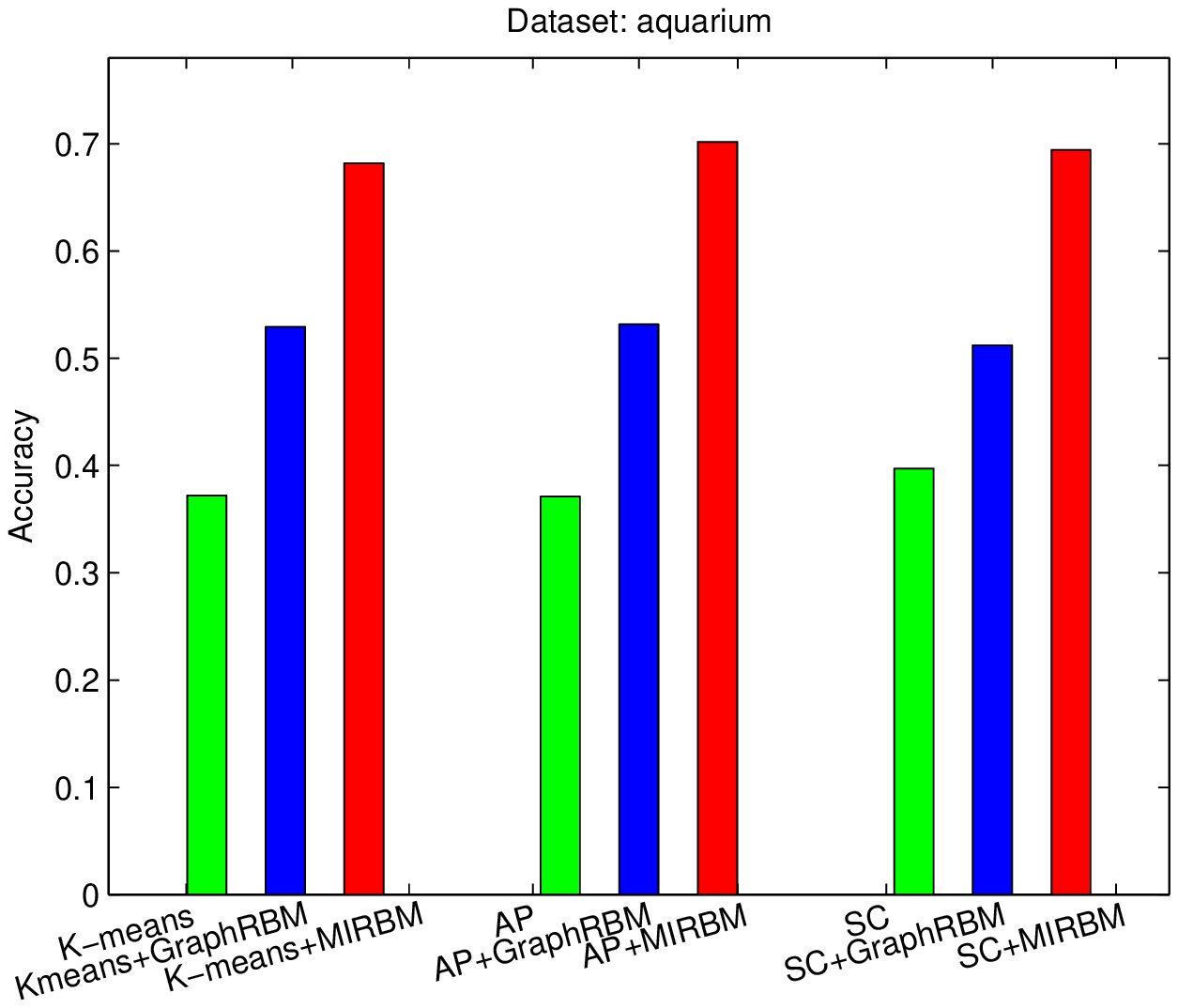}
     \includegraphics[scale=0.4]{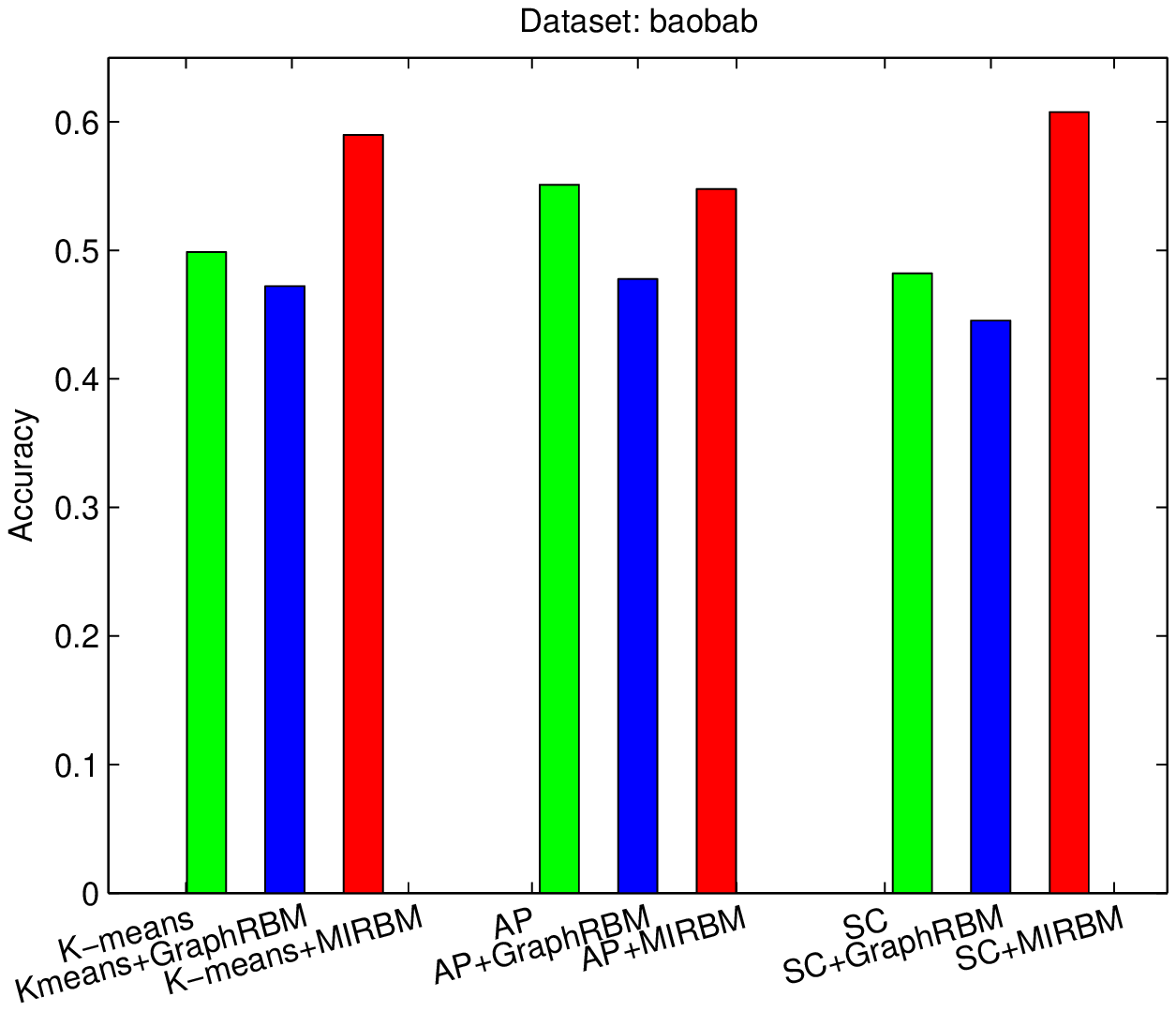}
    \includegraphics[scale=0.4]{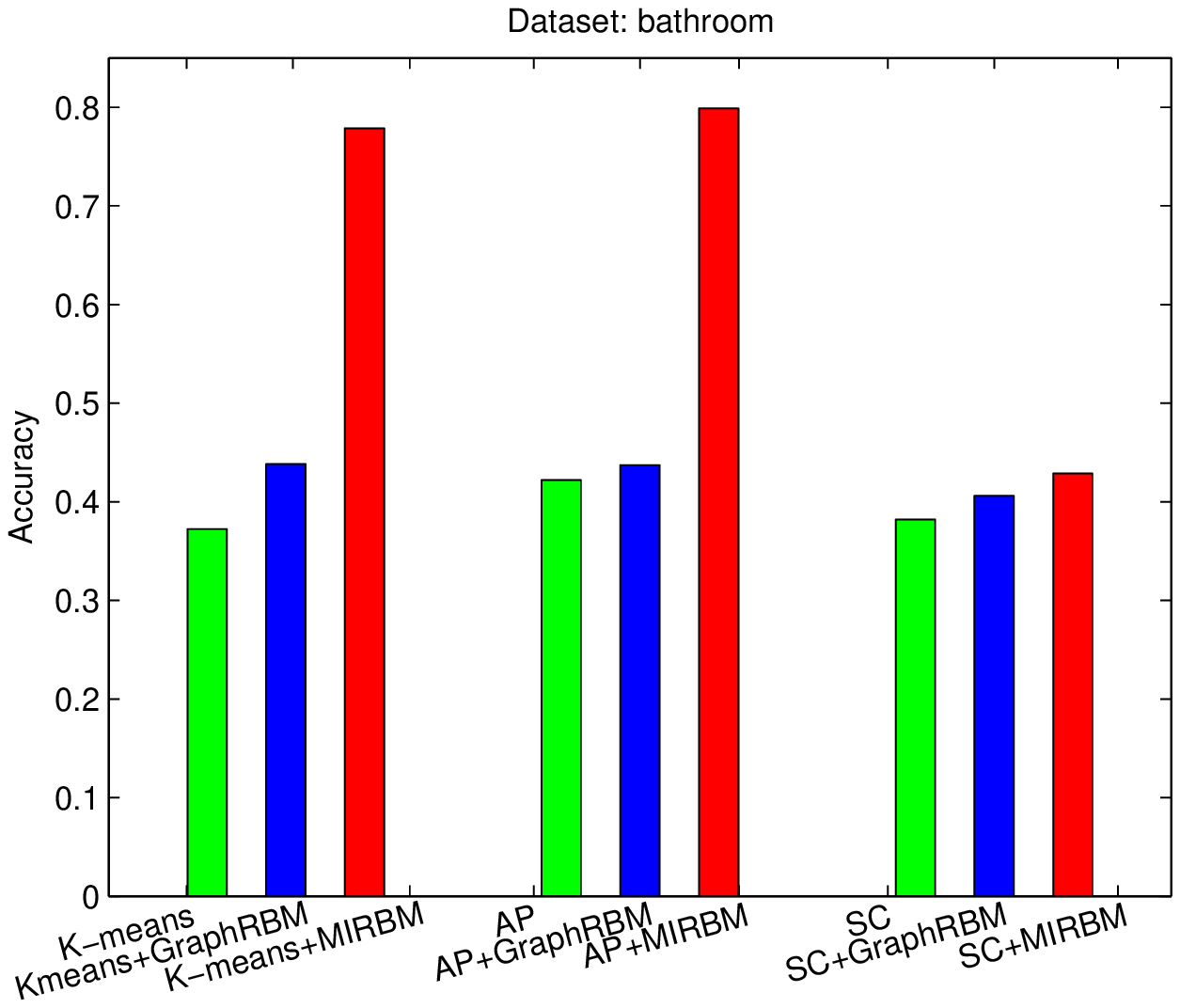}
     \includegraphics[scale=0.4]{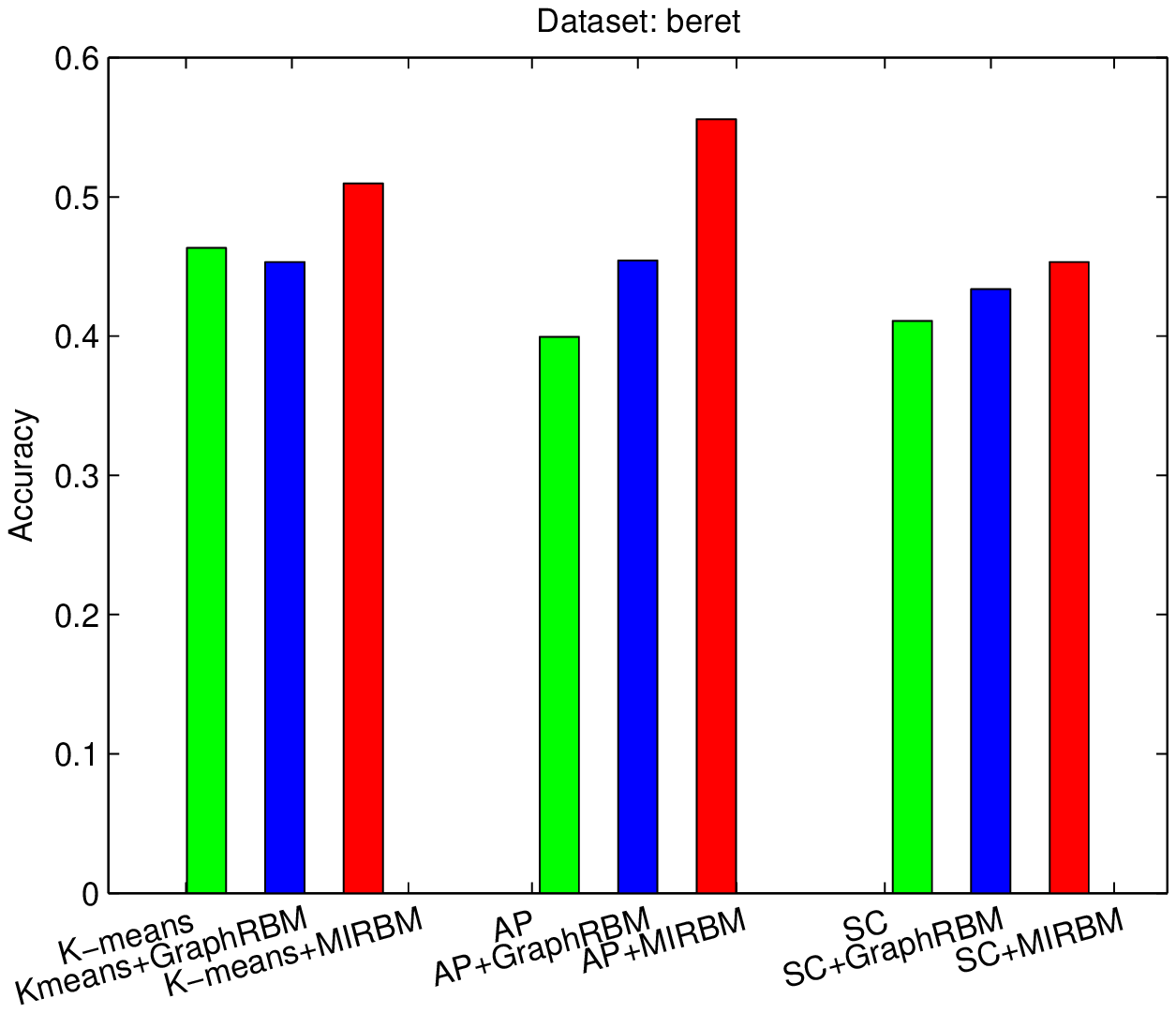}
     \includegraphics[scale=0.4]{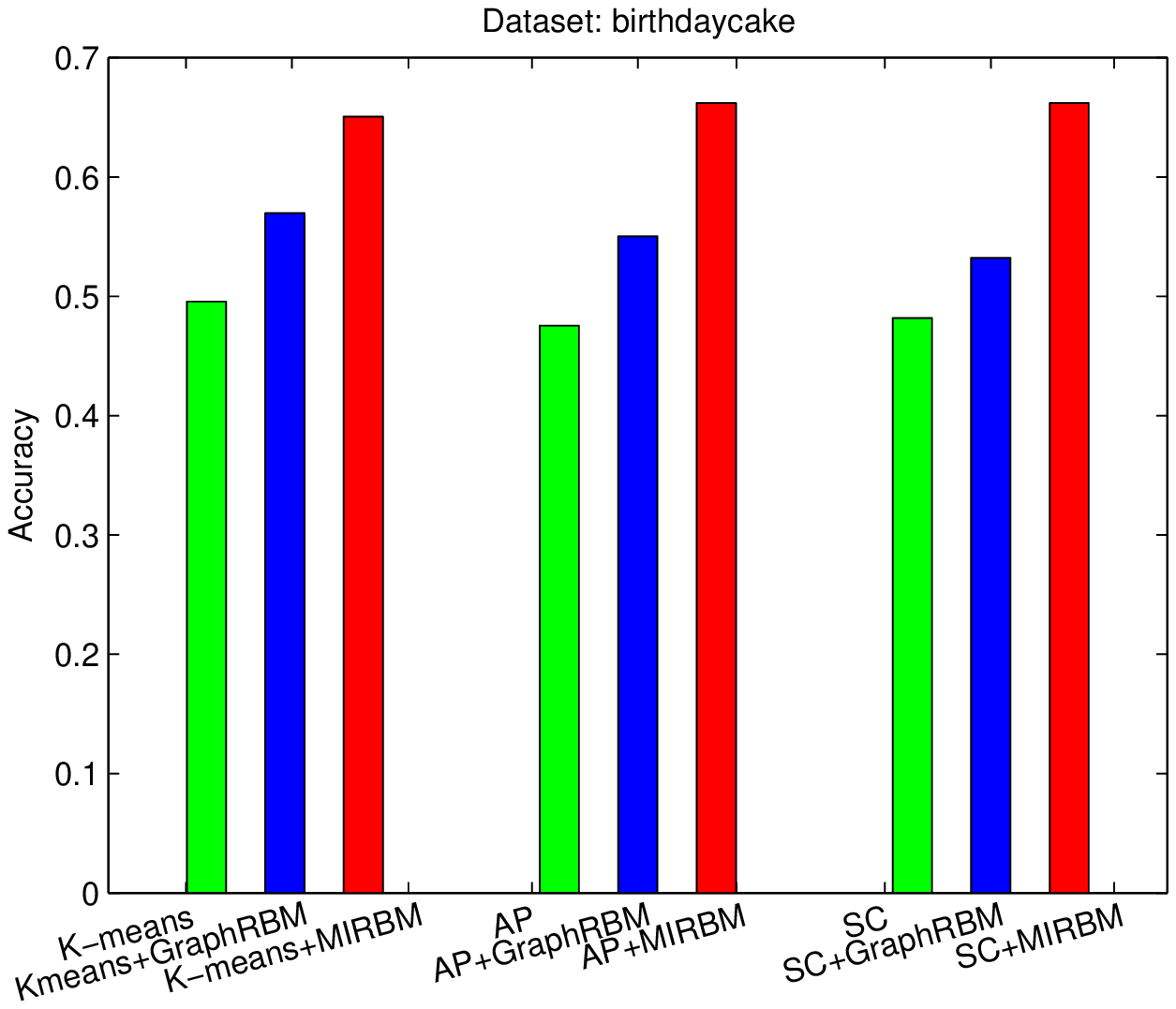}
    \includegraphics[scale=0.4]{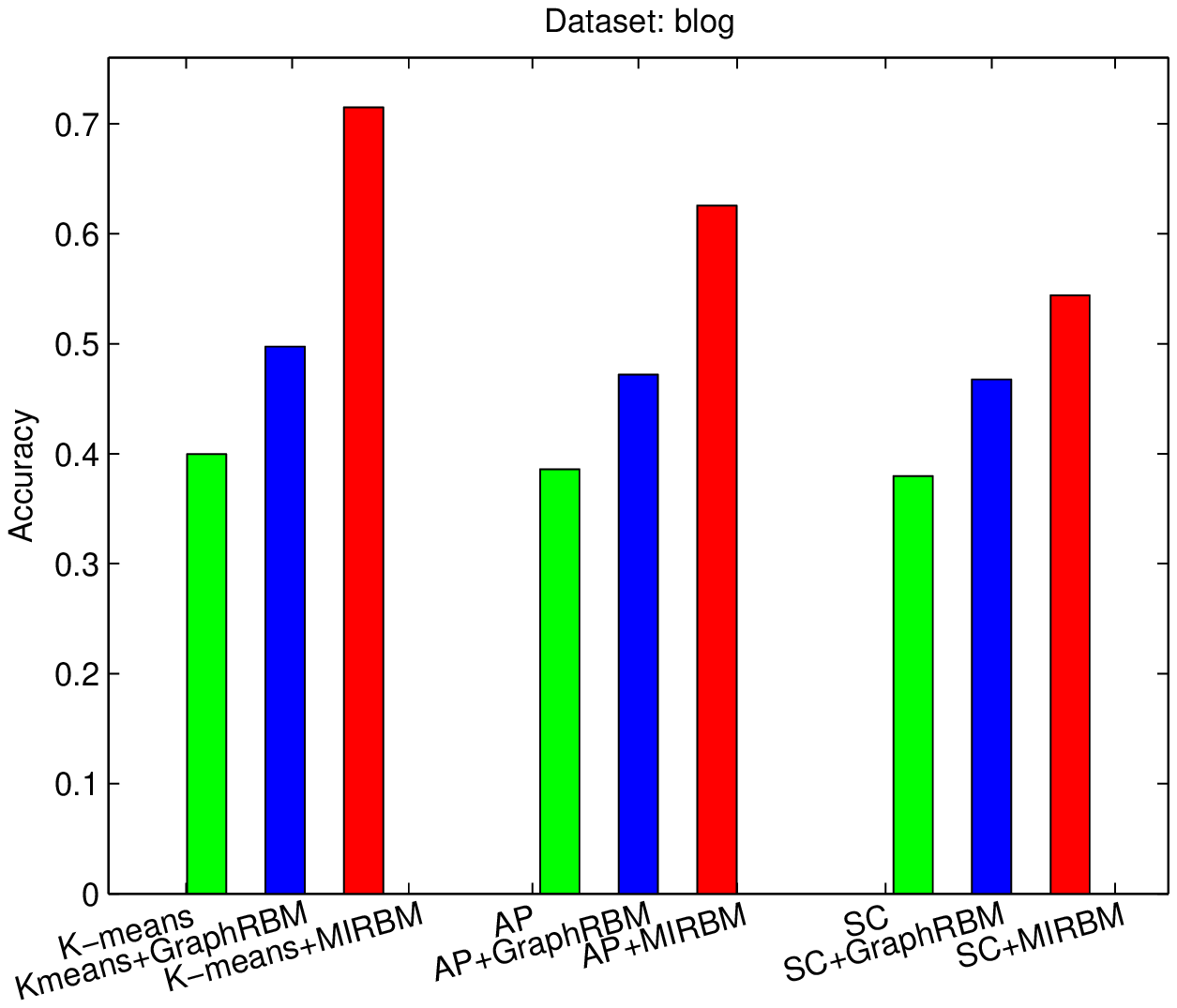}
     \includegraphics[scale=0.4]{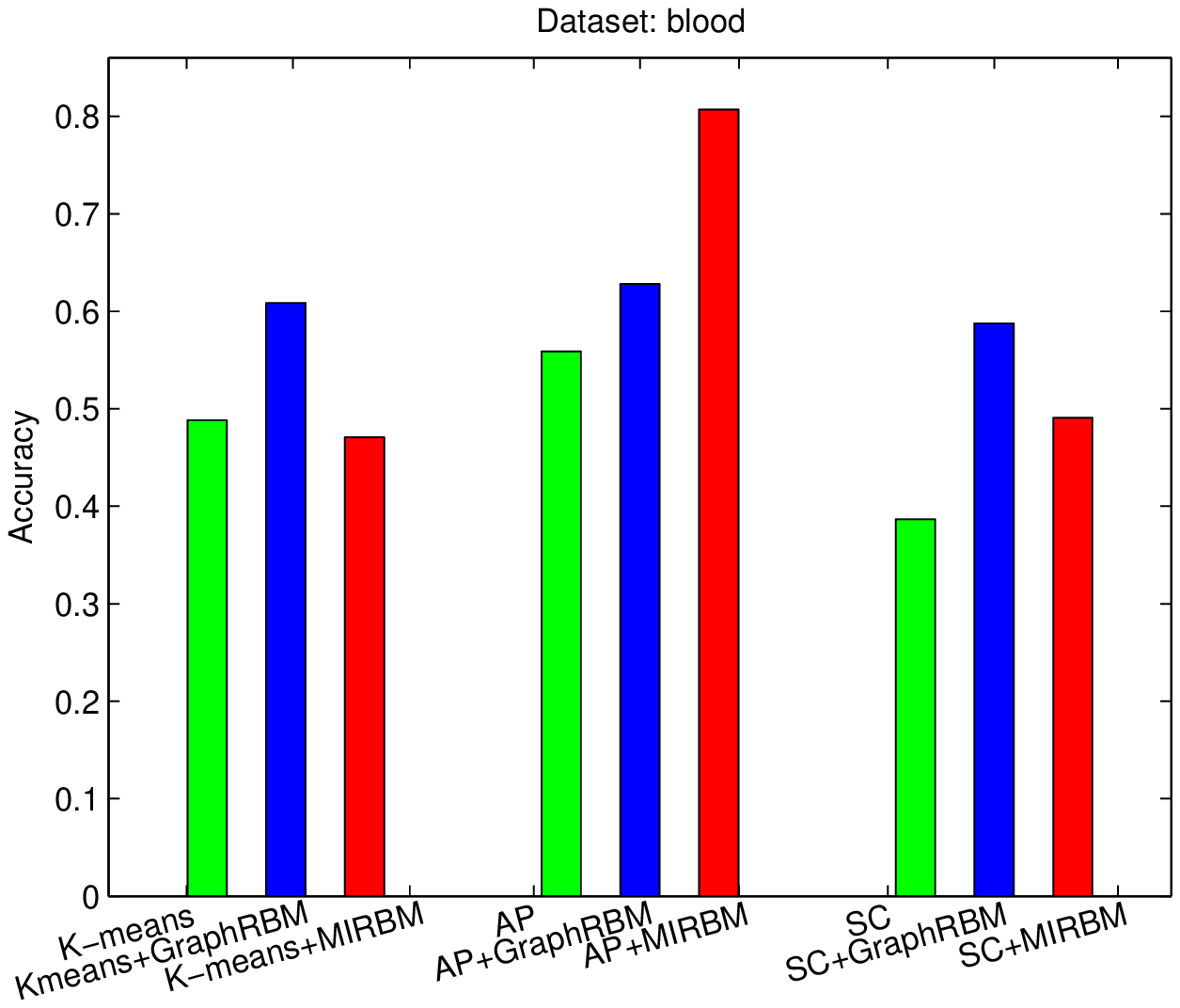}
     \includegraphics[scale=0.4]{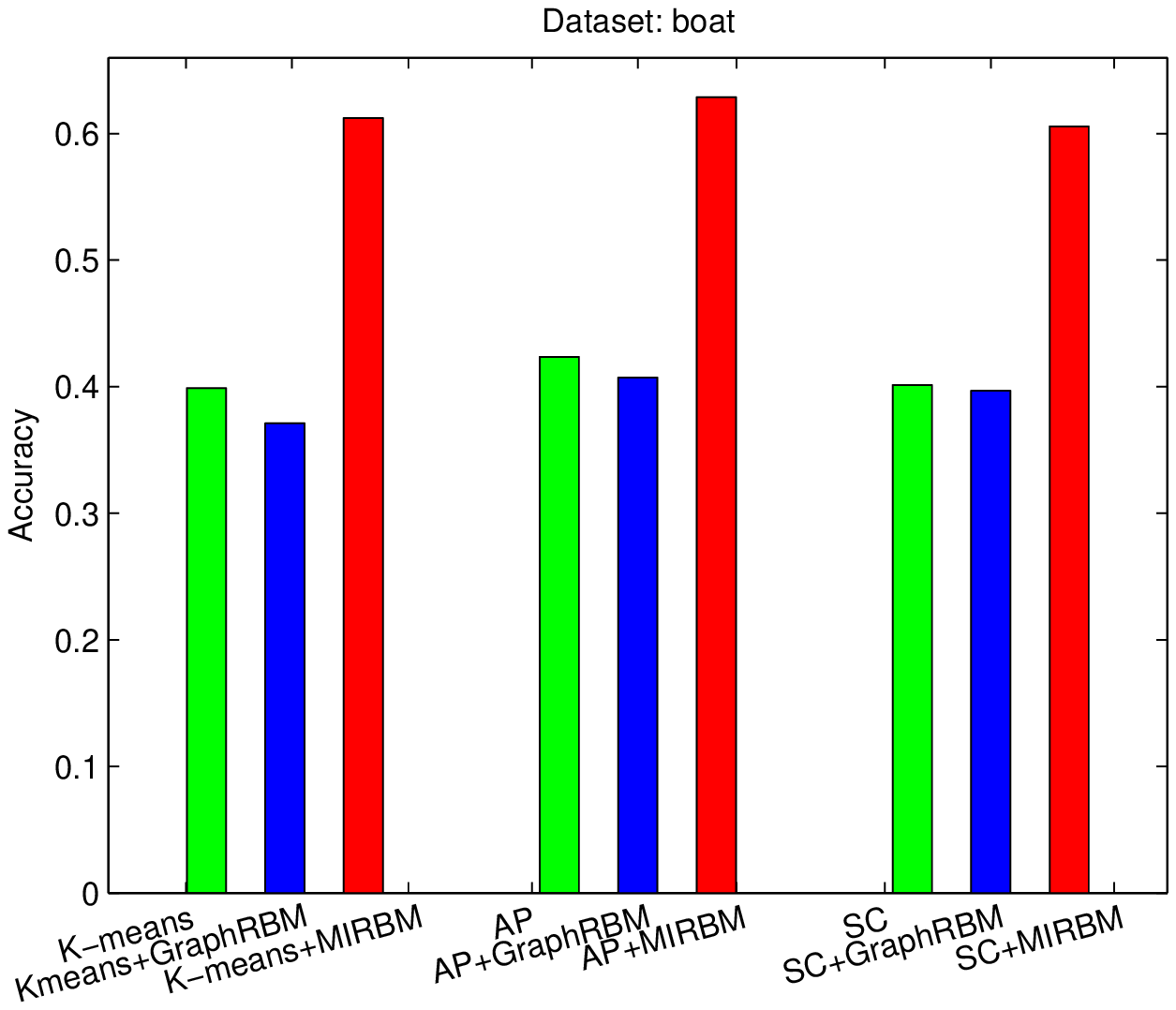}
    \includegraphics[scale=0.4]{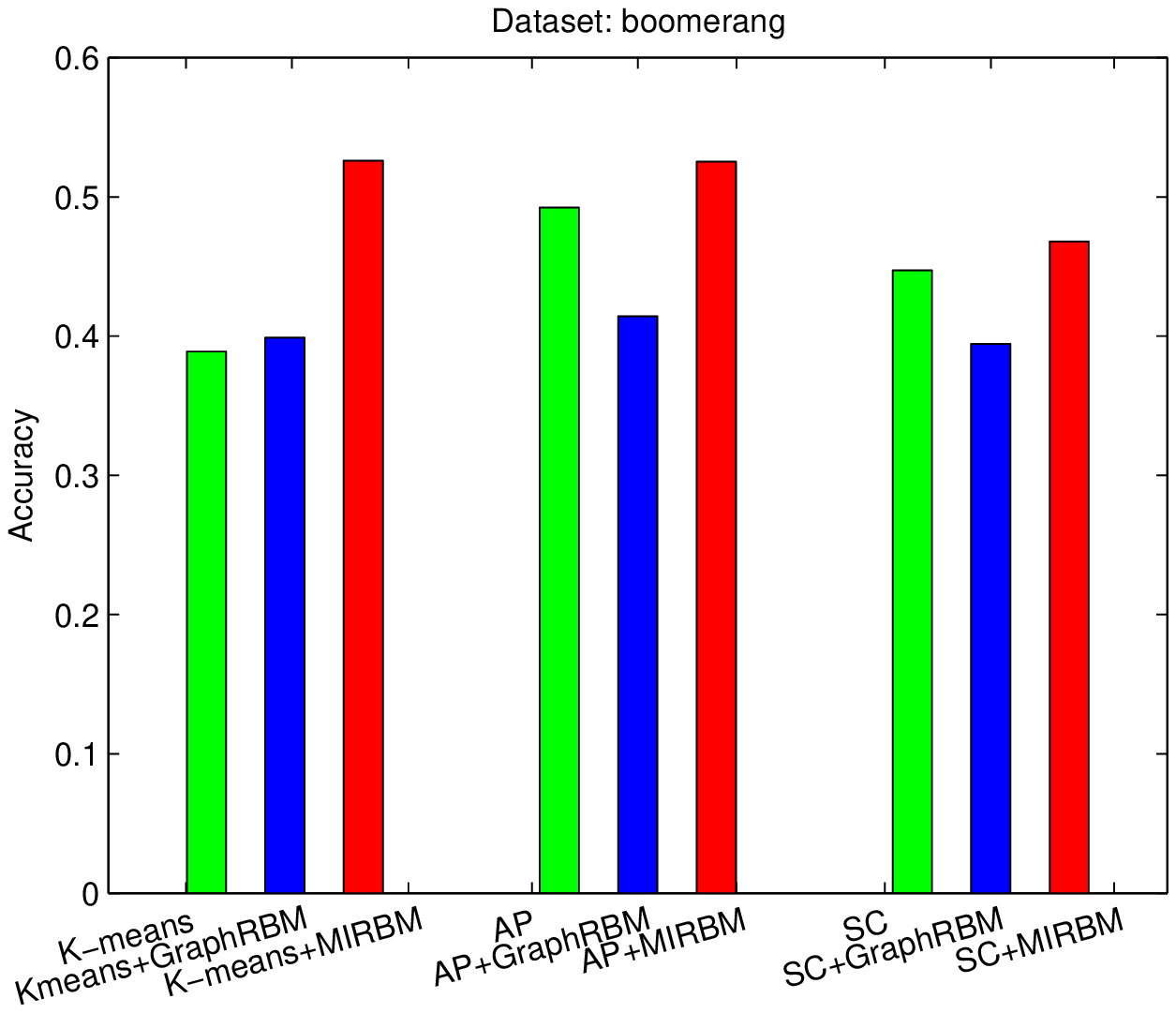}
     \includegraphics[scale=0.4]{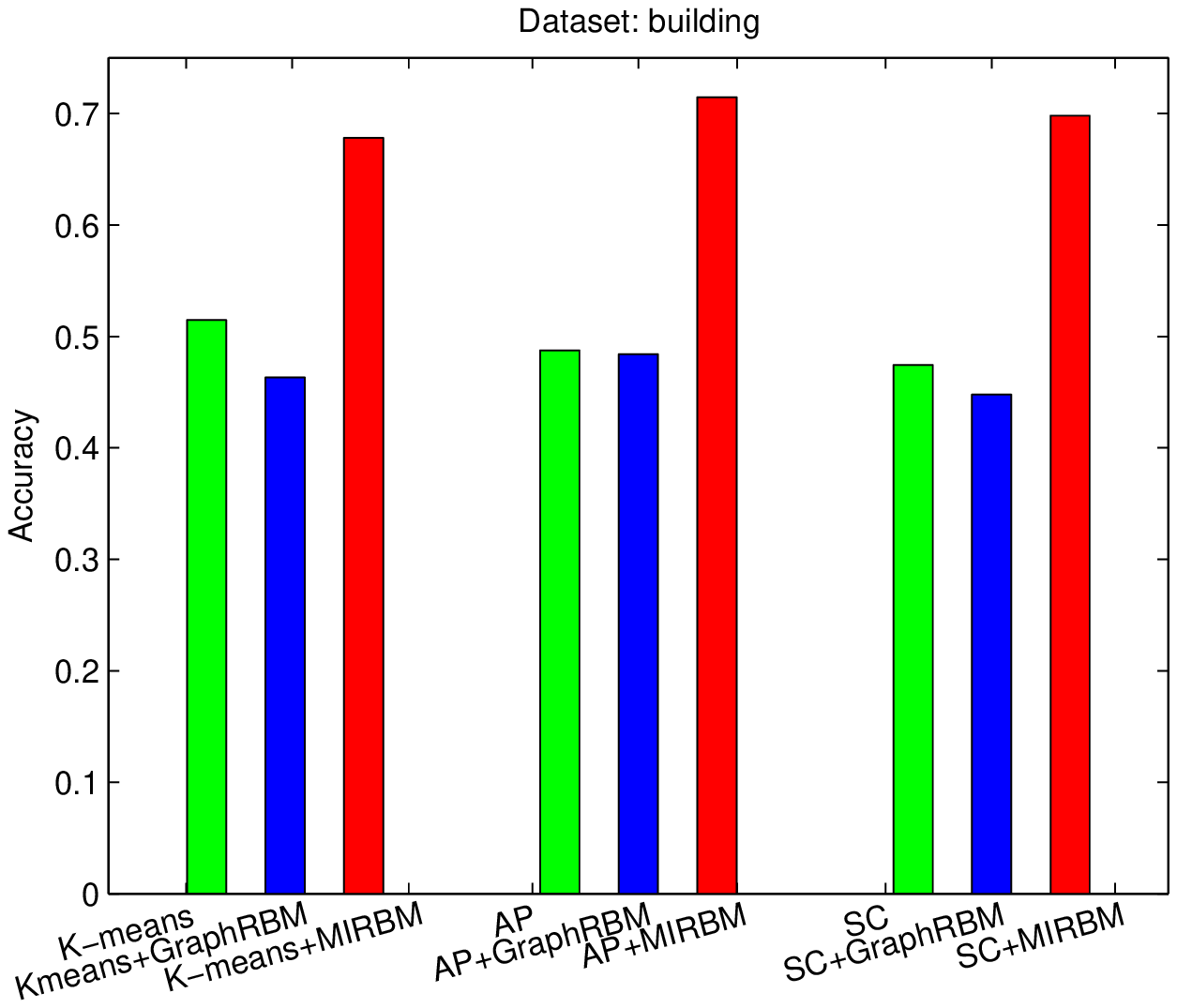}
    \includegraphics[scale=0.4]{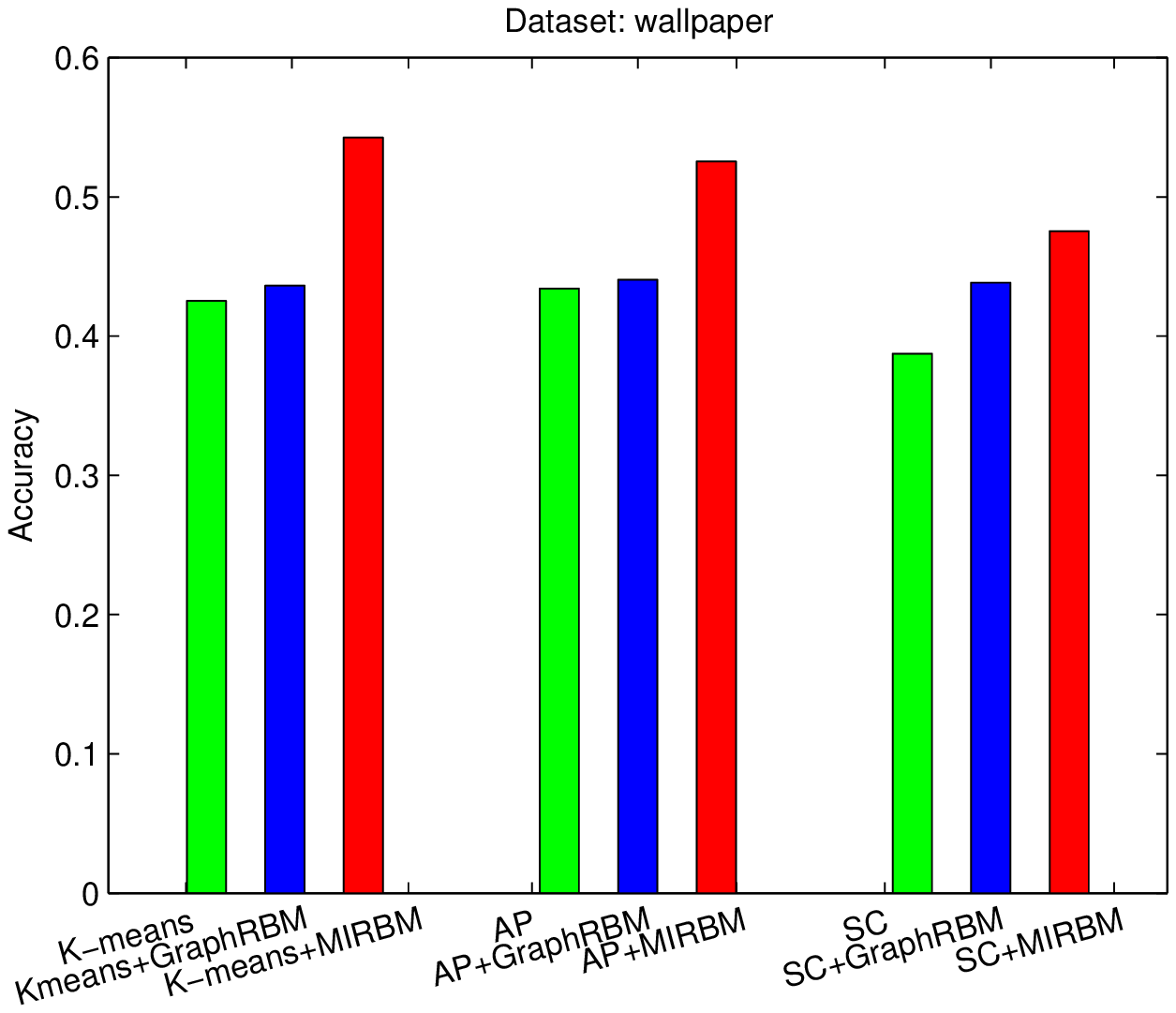}
     \includegraphics[scale=0.4]{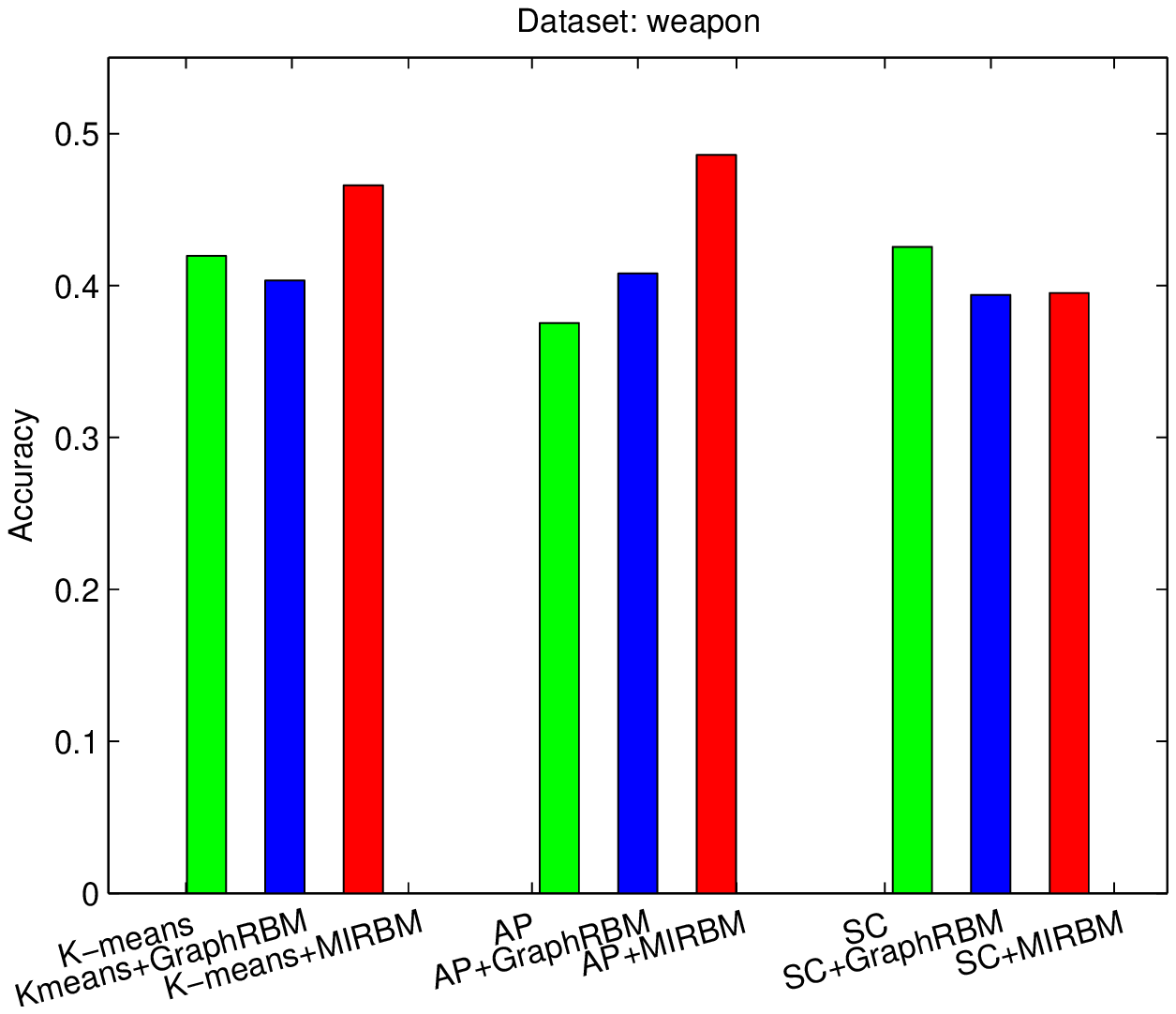}
    \includegraphics[scale=0.4]{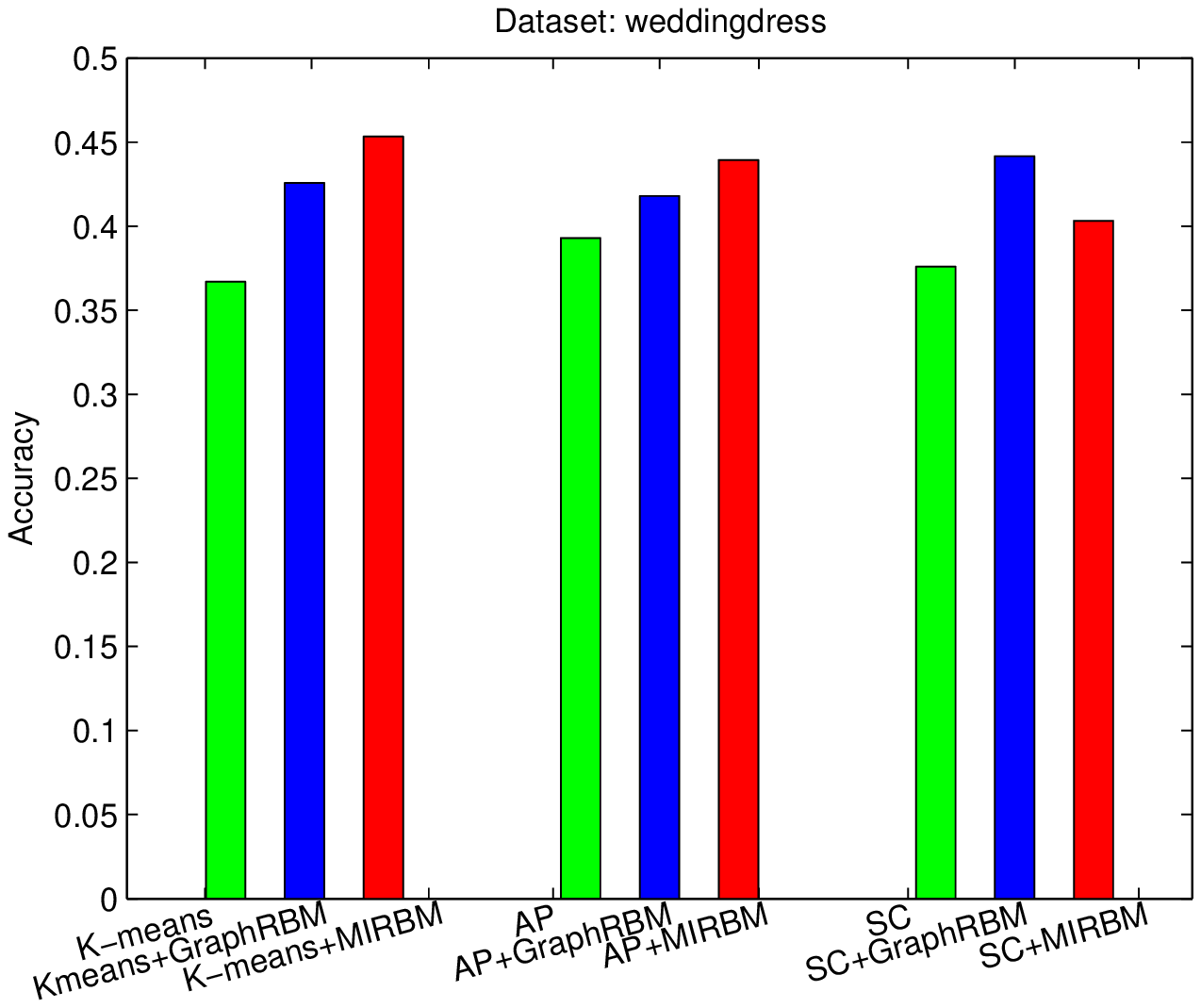}
\caption{Accuracy comparison between the proposed algorithms (K-means+MIRBM, AP+MIRBM and SC+MIRBM) and contrastive algorithms ($\eta=0.1$).
} \label{fig:1}
\end{figure*}
\begin{table*}
\begin{center}
\caption{The performances of the purity ($\eta=0.1$).}
\label{tab:results1} \scalebox{0.9}{
\begin{tabular}{|l|c|c|c|c|c|c|c|c|c|c|c|c|}
\hline
\textsf{ \bf{Dataset (No.)}} &  {K-means} & {AP} &{SC} &{K-means+GraphRBM}& {AP+GraphRBM} & {SC+GraphRBM}&{K-means+MIRBM}& {AP+MIRBM} & {SC+MIRBM}\\
\hline \textsf{alphabet} & {0.8395}& {	0.8453}& {	0.8447}& {	0.8425}& {	0.8474}& {	0.8442}& {	\textbf{0.8748}}& {	 0.8605}& {	0.8439} \\
\hline \textsf{americanflag} & {0.8375}& {	0.8212}& {	0.8320}&  {	0.8717}& {	0.8738}& {	0.8705}& {	0.8877}& {	 \textbf{0.8947}}& {	 0.8829}\\
\hline \textsf{aquarium} & {0.6961}& {	0.6983}& {	0.6952}& {	0.7074}& {	0.7125}& {	0.7050}& {	0.7148}& {	 \textbf{0.7218}}& {	 0.7193} \\
\hline \textsf{baobab} & {0.7676}& {	0.7528}& {	0.7659}&  {	0.7955}& {	\textbf{0.8003}}& {	0.7955}& {	0.7842}& {	 0.7644}& {	0.7909} \\
\hline \textsf{bathroom} & {0.5782}& {	0.5804}& {	0.5775}& {	0.5586}& {	0.5643}& {	0.5595}& {	0.5747}& {	 \textbf{0.5817}}& {	 0.5801}\\
\hline \textsf{beret} & {0.6942}& {	0.7095}& {	0.7032}& {	0.7132}& {	0.7182}& {	0.7111}& {	0.7135}& {	0.7212}& {	 \textbf{0.7232}}\\
\hline \textsf{birthdaycake} & {0.7342}& {	0.7694}& {	0.7420}& {	0.7931}& {	\textbf{0.7977}}& {	0.7901}& {	0.7093}& {	0.7314}& {	 0.7546}\\
\hline \textsf{blog} & {0.6723}& {	0.6684}& {	0.6749}& {	0.6697}& {	0.6740}& {	0.6704}& {	0.6754}& {	0.6789}& {	 \textbf{0.6811}} \\
\hline \textsf{blood} & {0.4615}& {	0.4608}& {	0.4606}&  {	0.4611}& {	0.4662}& {	0.4604}& {	0.4426}& {	\textbf{0.4682}}& {	0.4496}\\
\hline \textsf{boat} & {0.8102}& {	0.8260}& {	0.8018}& {	0.8109}& {	0.8212}& {	0.8109}& {	0.8231}& {	\textbf{0.8299}}& {	 0.8296} \\
\hline \textsf{boomerang} &  {0.9027}& {	0.8760}& {	0.8825}& {	0.9038}& {	0.9056}& {	0.8989}& {	0.8954}& {	\textbf{0.9041}}& {	 0.8911}\\
\hline \textsf{building} & {0.6317}& {	0.6637}& {	0.6452}& {	0.6872}& {	0.6924}& {	0.6871}& {	0.6956}& {	 \textbf{0.7024}}& {	 0.7020}\\
\hline \textsf{wallpaper} & {0.7797}& {	0.7829}& {	0.7770}& {	0.7752}& {	0.7800}& {	0.7755}& {	0.7811}& {	 \textbf{0.7873}}& {	 0.7836}\\
\hline \textsf{weapon} & {0.9428}& {	0.9364}& {	0.9165}& {	0.9415}& {	0.9465}& {	0.9397}& {	0.9478}& {	 \textbf{0.9548}}& {	 0.9345}\\
\hline \textsf{weddingdress} & {0.8921}& {	0.8845}& {	0.8868}&  {	0.8738}& {	0.8760}& {	0.8715}& {	0.8848}& {	\textbf{0.8935}}& {	 0.8889}\\
\hline
\hline  {Average} & {0.7483}& {	0.7509}& {	0.7465}&  {	0.7578}& {	0.7625}& {	0.7568}& {	\textbf{0.7590}}& {	\textbf{0.7653}}& {	 \textbf{0.7607}}\\
\hline
\end{tabular}}
\end{center}
\end{table*}
\begin{table*}
\begin{center}
\caption{The performances of the FMI ($\eta=0.1$).}
\label{tab:results1} \scalebox{0.9}{
\begin{tabular}{|l|c|c|c|c|c|c|c|c|c|c|c|c|}
\hline
\textsf{ \bf{Dataset (No.)}} &  {K-means} & {AP} &{SC} &{K-means+GraphRBM}& {AP+GraphRBM} & {SC+GraphRBM}&{K-means+MIRBM}& {AP+MIRBM} & {SC+MIRBM}\\
\hline \textsf{alphabet} &  {0.3791}& {	0.3808}& {	0.3817}&{	0.3746}& {	0.3753}& {	0.3738}& {	0.5769}& {	 \textbf{0.5855}}& {	 0.4463} \\
\hline \textsf{americanflag} &{0.4164}& {	0.4101}& {	0.4046}&  {	0.4306}& {	0.4431}& {	0.4181}& {	0.5843}& {	 \textbf{0.6443}}& {	 0.6101}\\
\hline \textsf{aquarium} & {0.4412}& {	0.4323}& {	0.4323}&  {	0.5133}& {	0.5147}& {	0.4920}& {	0.6770}& {	 \textbf{0.7370}}& {	 0.7282}\\
\hline \textsf{baobab} & {0.4659}& {	0.4931}& {	0.4580}&  {	0.4844}& {	0.4890}& {	0.4638}& {	0.5451}& {	0.5055}& {	 \textbf{0.6030}}\\
\hline \textsf{bathroom} & {0.4729}& {	0.4849}& {	0.4726}& {	0.4922}& {	0.4910}& {	0.4904}& {	0.7514}& {	 \textbf{0.8114}}& {	 0.4996}  \\
\hline \textsf{beret} & {0.4746}& {	0.4337}& {	0.4353}&  {	0.4631}& {	0.4643}& {	0.4575}& {	0.4747}& {\textbf{	 0.5654}}& {	0.4783}\\
\hline \textsf{birthdaycake} & {0.4612}& {	0.4288}& {	0.4513}&  {	0.5434}& {	0.5183}& {	0.5137}& {	0.6069}& {	\textbf{0.6618}}& {	 0.6594}  \\
\hline \textsf{blog} & {0.4502}& {	0.4477}& {	0.4453}& {	0.5031}& {	0.4842}& {	0.4839}& {	\textbf{0.7003}}& {	0.6256}& {	0.5472} \\
\hline \textsf{blood} & {0.5290}& {	0.5769}& {	0.4923}& {	0.6117}& {	0.6282}& {	0.5946}& {	0.5258}& {	\textbf{0.8166}}& {	0.5858} \\
\hline \textsf{boat} &{0.4212}& {	0.4212}& {	0.4181}&  {	0.4053}& {	0.4204}& {	0.4104}& {	0.6234}& {	 \textbf{0.6772}}& {	0.6398} \\
\hline \textsf{boomerang} & {0.3719}& {	0.4143}& {	0.3884}&  {	0.3926}& {	0.4026}& {	0.3911}& {	0.5089}& {	\textbf{0.5270}}& {	 0.4307} \\
\hline \textsf{building} & {0.4876}& {	0.5060}& {	0.4669}&  {	0.4711}& {	0.4861}& {	0.4696}& {	0.6657}& {	 \textbf{0.7456}}& {	 0.7255}\\
\hline \textsf{wallpaper} & {0.4153}& {	0.4154}& {	0.4044}& {	0.4737}& {	0.4845}& {	0.4671}& {	\textbf{0.6182}}& {	 0.5430}& {	0.4921}\\
\hline \textsf{weapon} & {0.4366}& {	0.3665}& {	0.3695}&  {	0.3863}& {	0.4058}& {	0.3818}& {	0.5463}& {	 \textbf{0.6063}}& {	 0.4728}\\
\hline \textsf{weddingdress} &{0.3702}& {	0.3732}& {	0.3721}&  {	0.3914}& {	0.3802}& {	0.3890}& {	0.4555}& {	 \textbf{0.5026}}& {	 0.4036}\\
\hline
\hline  {Average} & {0.4382}& {	0.4359}& {	0.4238}&  {	0.4619}& {	0.4653}& {	0.4529}& {	\textbf{0.5977}}& {	\textbf{0.6322}}& {	 \textbf{0.5482}}\\
\hline
\end{tabular}}
\end{center}
\end{table*}
\begin{table*}
\begin{center}
\caption{The performances of the Jac ($\eta=0.1$).}
\label{tab:results1} \scalebox{0.9}{
\begin{tabular}{|l|c|c|c|c|c|c|c|c|c|c|c|c|}
\hline
\textsf{ \bf{Dataset (No.)}} &  {K-means} & {AP} &{SC} & {K-means+GraphRBM}& {AP+GraphRBM} & {SC+GraphRBM}&{K-means+MIRBM}& {AP+MIRBM} & {SC+MIRBM}\\
\hline \textsf{alphabet} &  {0.2330}& {	0.2341}& {	0.2345}& {	0.2297}& {	0.2302}& {	0.2287}& {	0.3557}& {	 \textbf{0.3852}}& {	 0.2871} \\
\hline \textsf{americanflag} &{0.2612}& {	0.2578}& {	0.2521}&  {	0.2739}& {	0.2833}& {	0.2643}& {	0.3652}& {	 \textbf{0.4152}}& {	 0.3923}\\
\hline \textsf{aquarium} & {0.2767}& {	0.2687}& {	0.2665}& {	0.3437}& {	0.3451}& {	0.3234}& {	0.4932}& {	 \textbf{0.5432}}& {	 0.5350}\\
\hline \textsf{baobab} &{0.3008}& {	0.3252}& {	0.2932}& {	0.3195}& {	0.3235}& {	0.3018}& {	0.3752}& {	0.3363}& {	 \textbf{0.4240}}   \\
\hline \textsf{bathroom} & {0.2885}& {	0.3024}& {	0.2878}&  {	0.3051}& {	0.3041}& {	0.3036}& {	0.6083}& {	 \textbf{0.6583}}& {	 0.3181} \\
\hline \textsf{beret} &{0.3031}& {	0.2695}& {	0.2686}& {	0.2968}& {	0.2979}& {	0.2908}& {	0.3148}& {	 \textbf{0.3926}}& {	0.3119} \\
\hline \textsf{birthdaycake} & {0.2957}& {	0.2691}& {	0.2873}& {	0.3713}& {	0.3496}& {	0.3454}& {	0.4332}& {	\textbf{0.4760}}& {	 0.4660} \\
\hline \textsf{blog} & {0.2773}& {	0.2767}& {	0.2732}& {	0.3299}& {	0.3111}& {	0.3114}& {	\textbf{0.5280}}& {	0.4522}& {	 0.3761} \\
\hline \textsf{blood} & {0.3403}& {	0.3940}& {	0.2997}&  {	0.4319}& {	0.4516}& {	0.4118}& {	0.3544}& {	\textbf{0.6787}}& {	0.4044}\\
\hline \textsf{boat} & {0.2627}& {	0.2650}& {	0.2598}& {	0.2511}& {	0.2645}& {	0.2552}& {	0.4171}& {	 \textbf{0.4631}}& {	0.4393}\\
\hline \textsf{boomerang} & {0.2276}& {	0.2607}& {	0.2399}& {	0.2442}& {	0.2520}& {	0.2431}& {	0.3227}& {	 \textbf{0.3460}}& {	 0.2744}\\
\hline \textsf{building} & {0.3126}& {	0.3303}& {	0.2924}&  {	0.3019}& {	0.3168}& {	0.3007}& {	0.4891}& {	 \textbf{0.5567}}& {	 0.5390}\\
\hline \textsf{wallpaper} & {0.2595}& {	0.2599}& {	0.2504}& {	0.3101}& {	0.3190}& {	0.3046}& {	\textbf{0.4100}}& {	 0.3661}& {	0.3251}\\
\hline \textsf{weapon} & {0.2747}& {	0.2243}& {	0.2263}&{	0.2391}& {	0.2532}& {	0.2358}& {	0.3176}& {	 \textbf{0.3676}}& {	 0.2955}\\
\hline \textsf{weddingdress} &{0.2263}& {	0.2285}& {	0.2276}& {	0.2430}& {	0.2338}& {	0.2410}& {	0.2838}& {	 \textbf{0.3228}}& {	 0.2528}\\
\hline
\hline  {Average} & {0.2748}& {	0.2749}& {	0.2618}&  {	0.2989}& {	0.3018}& {	0.2905}& {	\textbf{0.4098}}& {\textbf{	0.4463}}& {	 \textbf{0.3702}}\\
\hline
\end{tabular}}
\end{center}
\end{table*}
\begin{figure*}
 \centering
    \includegraphics[scale=0.4]{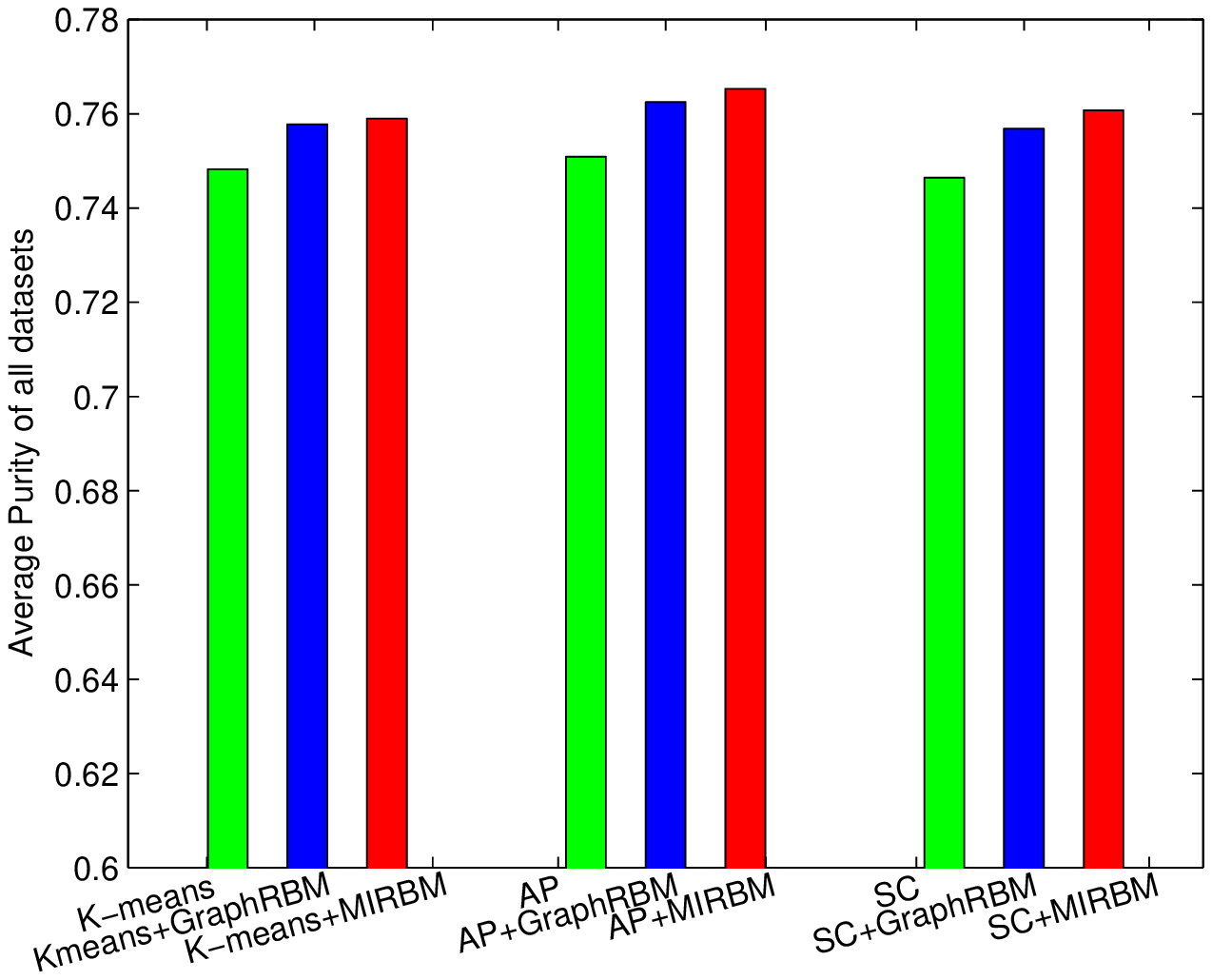}
     \includegraphics[scale=0.4]{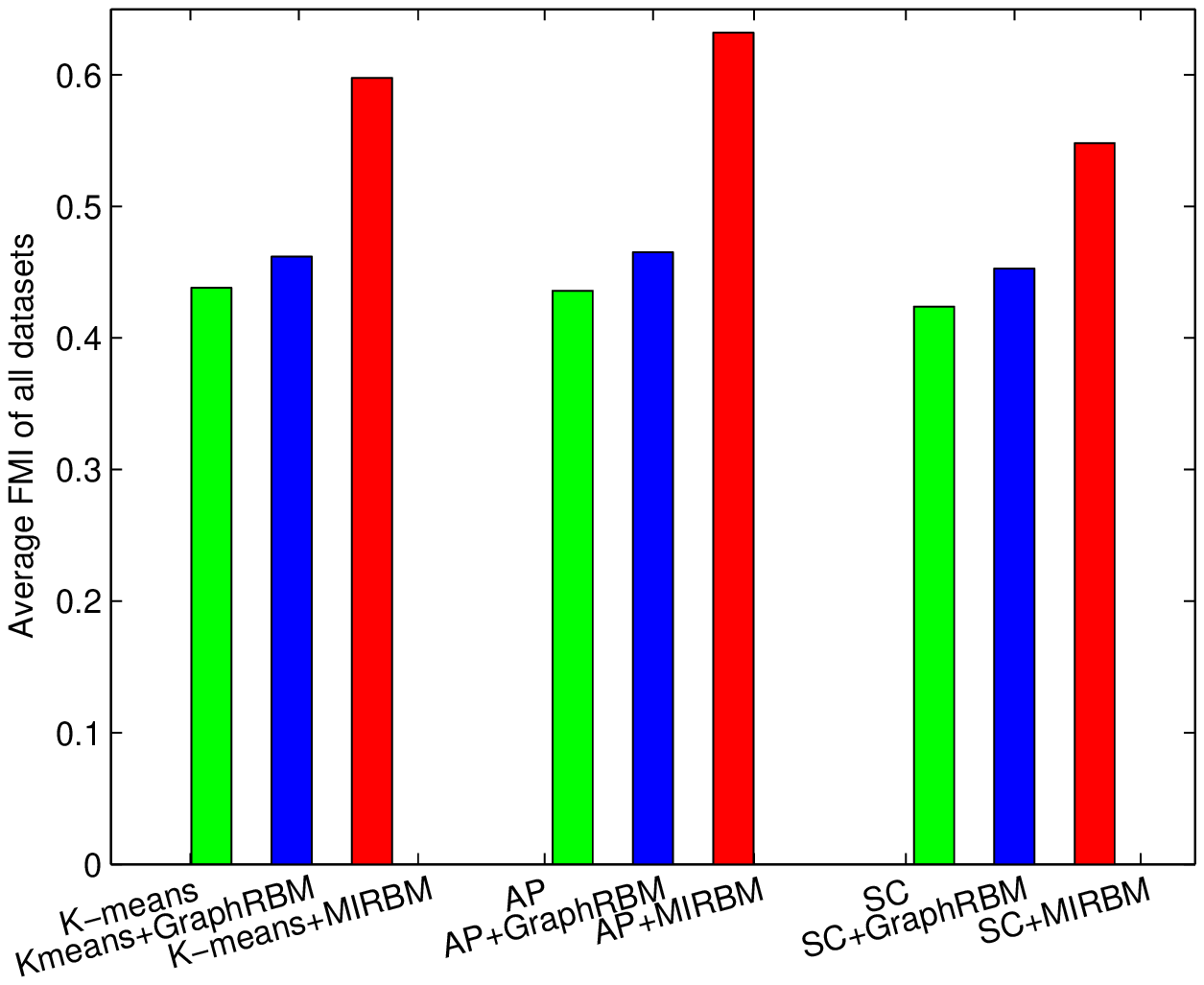}
 \includegraphics[scale=0.4]{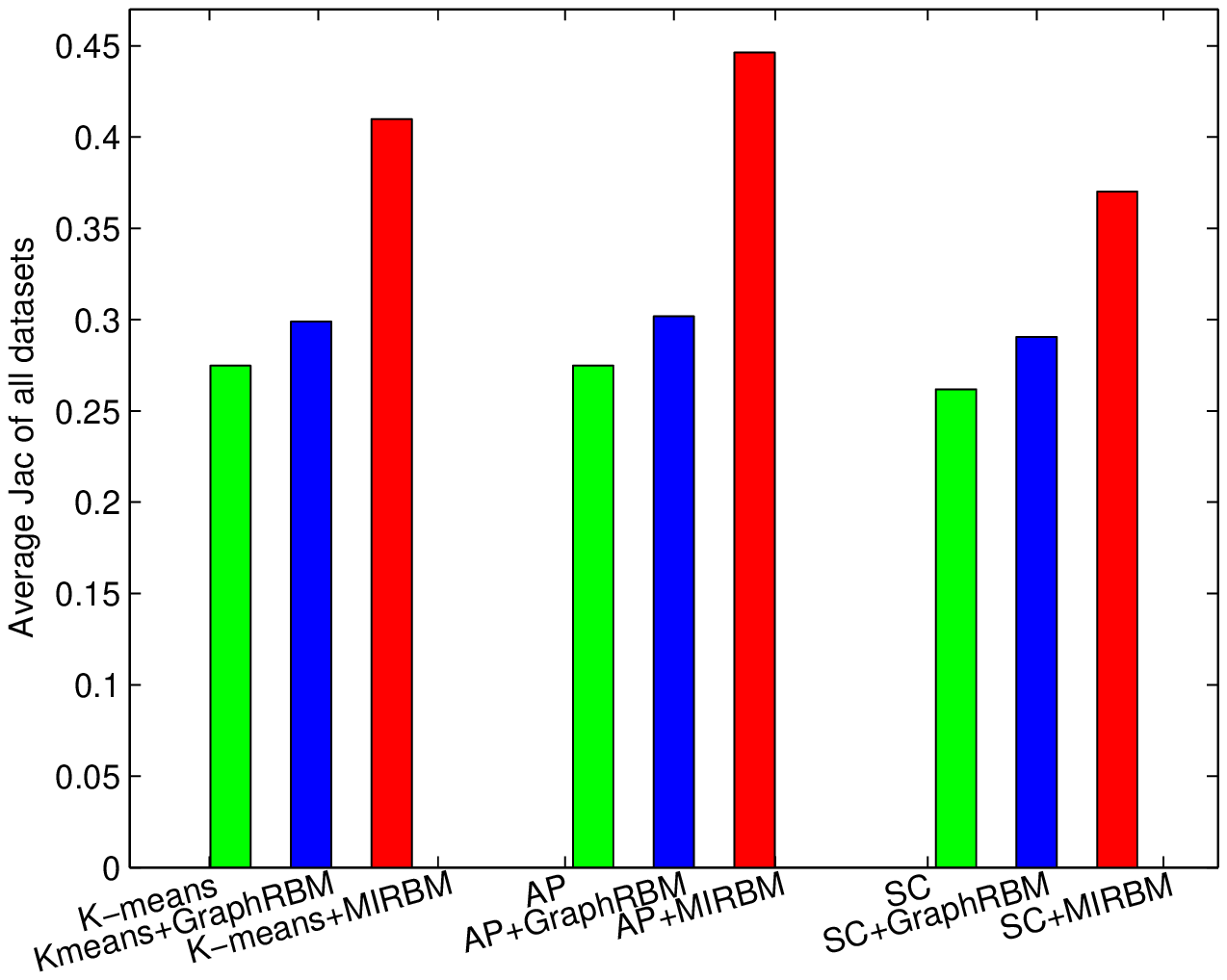}
\caption{The performances of average purity and FMI and Jac ($\eta=0.1$).}
\label{fig:1}
\end{figure*}
\subsection{Generalization Ability}
From Tables III, IV, V and VI, the K-means+MIRBM, AP+MIRBM and SC+MIRBM algorithms based on the proposed architecture show the best performance in their own groups with four evaluation metrics. With regard to the accuracy, FMI and Jac evaluation metrics, the K-means+MIRBM, AP+MIRBM and SC+MIRBM algorithms show the best performance than other groups algorithms. These mean that the hidden layer features of the proposed architecture are available for three different clustering algorithms. Hence, our architecture shows excellent generalization ability for unsupervised feature learning.
\subsection{Sensitivity Analysis of the Key Parameter $\eta$}
The parameter $\eta$ is an important scale coefficient in the proposed architecture. It has a direct impact on the role of auxiliary guidance LCP. Hence, we need analysis its sensitivity. The results of K-means+MIRBM, AP+MIRBM and SC+MIRBM algorithms based on our architecture are shown in Fig. 5. The accuracy, purity, FMI and Jac evaluation metrics show the best performance when the parameter $\eta$ is set to 0.1. The performance of accuracy decrease rapidly with the increase of parameter $\eta$ from 0.1 to 0.9. The performance of FMI and Jac decrease rapidly with the increase of parameter $\eta$ except for K-means+MIRBM algorithm. But, the purity is insensitive to other evaluating metrics with the increase of parameter $\eta$.
\begin{figure*}
 \centering
     \includegraphics[scale=0.45]{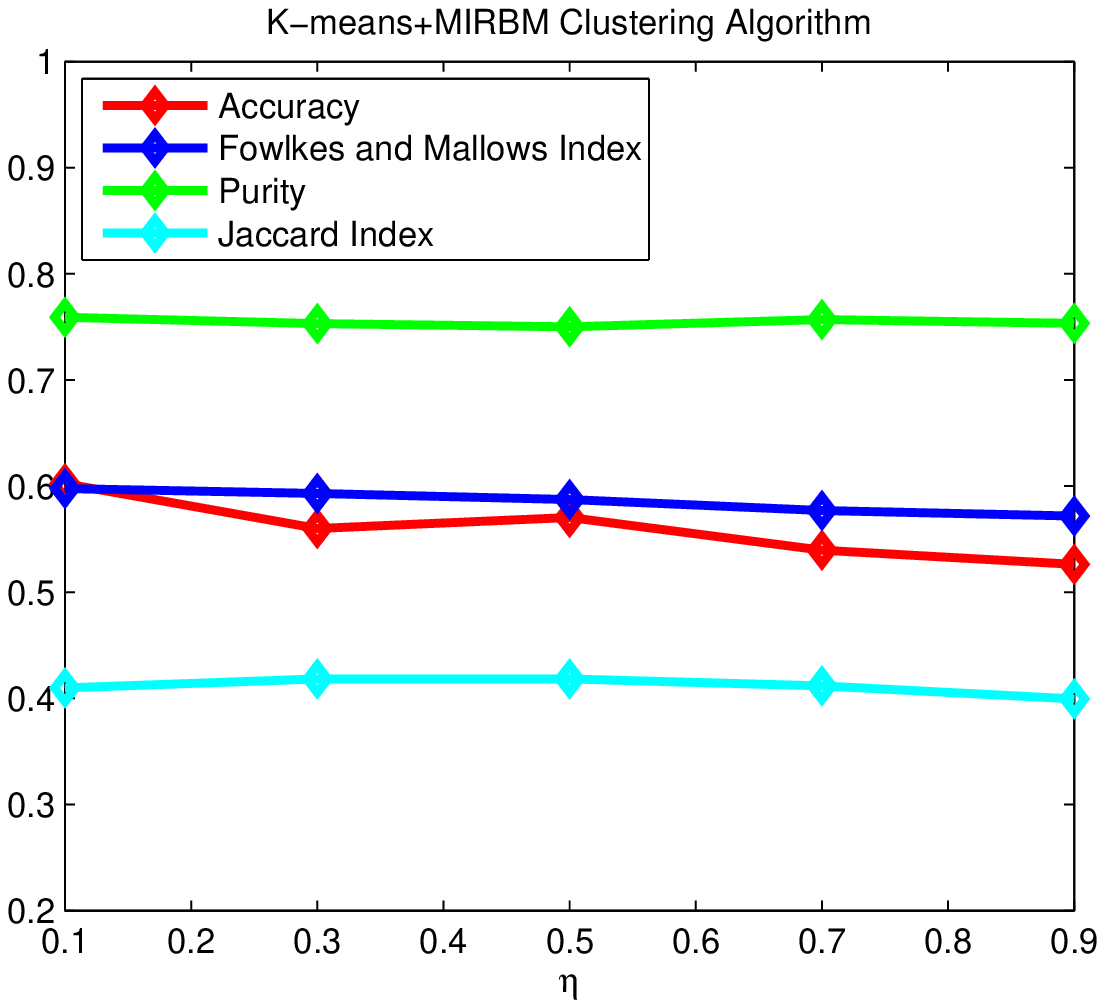}
     \includegraphics[scale=0.45]{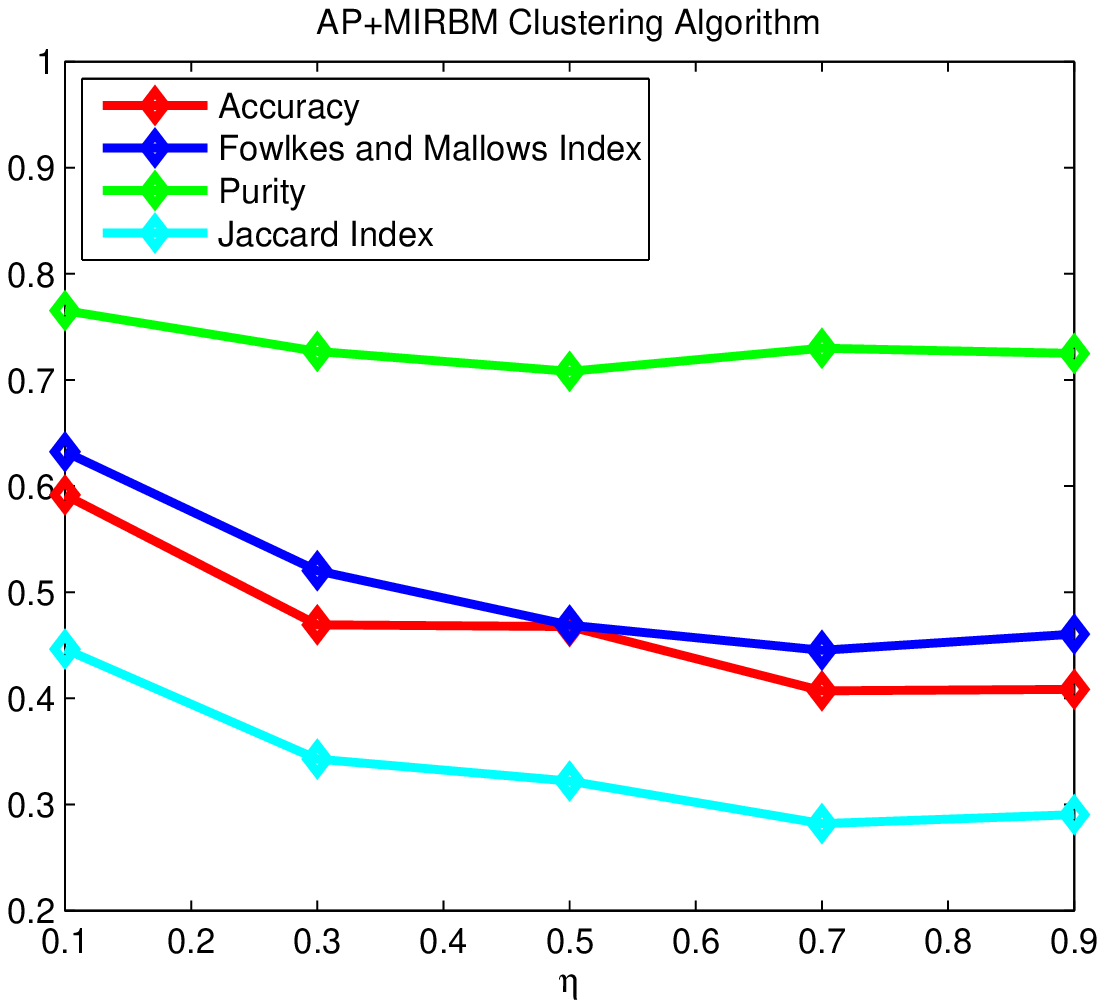}
     \includegraphics[scale=0.45]{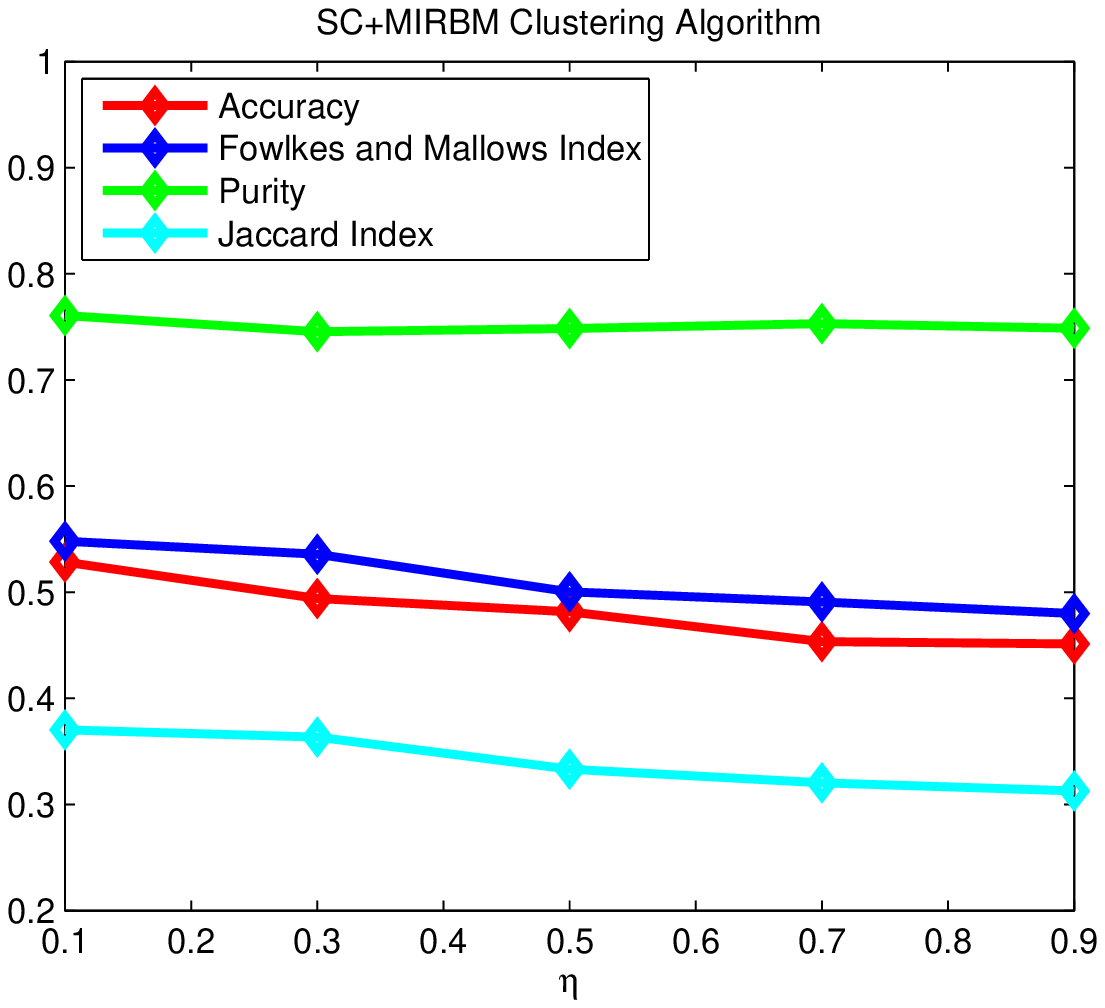}
\caption{Sensitivity analysis of the key parameters $\eta$ for the K-means+MIRBM, AP+MIRBM and SC+MIRBM algorithms based on the proposed architecture.
} \label{fig:1}
\end{figure*}
\section{Conclusions}
\indent We presented a novel unsupervised feature learning architecture for image clustering. The main motivation behind the proposed architecture was the need to guide feature distribution for clustering by unsupervised LCP in the training process. We compared the proposed architecture with state-of-the-art GraphRBM to reveal the feature representation capability. In our experimental datasets, the proposed architecture showed better performance than the GraphRBM. For an important scale coefficient $\eta$, we studied its sensitivity. When the parameter $\eta=0.1$, the accuracy, purity, FMI and Jac evaluation metrics showed the best performance. The performance of accuracy decreased rapidly with the increase of parameter $\eta$. The performance of FMI and Jac decreased rapidly with the increase of parameter $\eta$ except for K-means+MIRBM algorithm. But, the purity was insensitive to other evaluating metrics with the increase of parameter $\eta$ for K-means+MIRBM, AP+MIRBM and SC+MIRBM algorithms. Furthermore, we demonstrated that the proposed architecture has excellent generalization ability for different clustering algorithms.\\
\indent For future work, we would like to exploit more efficient unsupervised feature learning architecture for large-scale datasets. Deep architecture is also an interesting work in the future.
\section{Acknowledgement}
This work were partially supported by  the National Science Foundation of China (Nos. 61773324, 61573292 and 61976247).
\bibliography{rbm}
\bibliographystyle{IEEEtran}
\begin{IEEEbiography}[{\includegraphics [width=1in,height=1.25in,keepaspectratio]{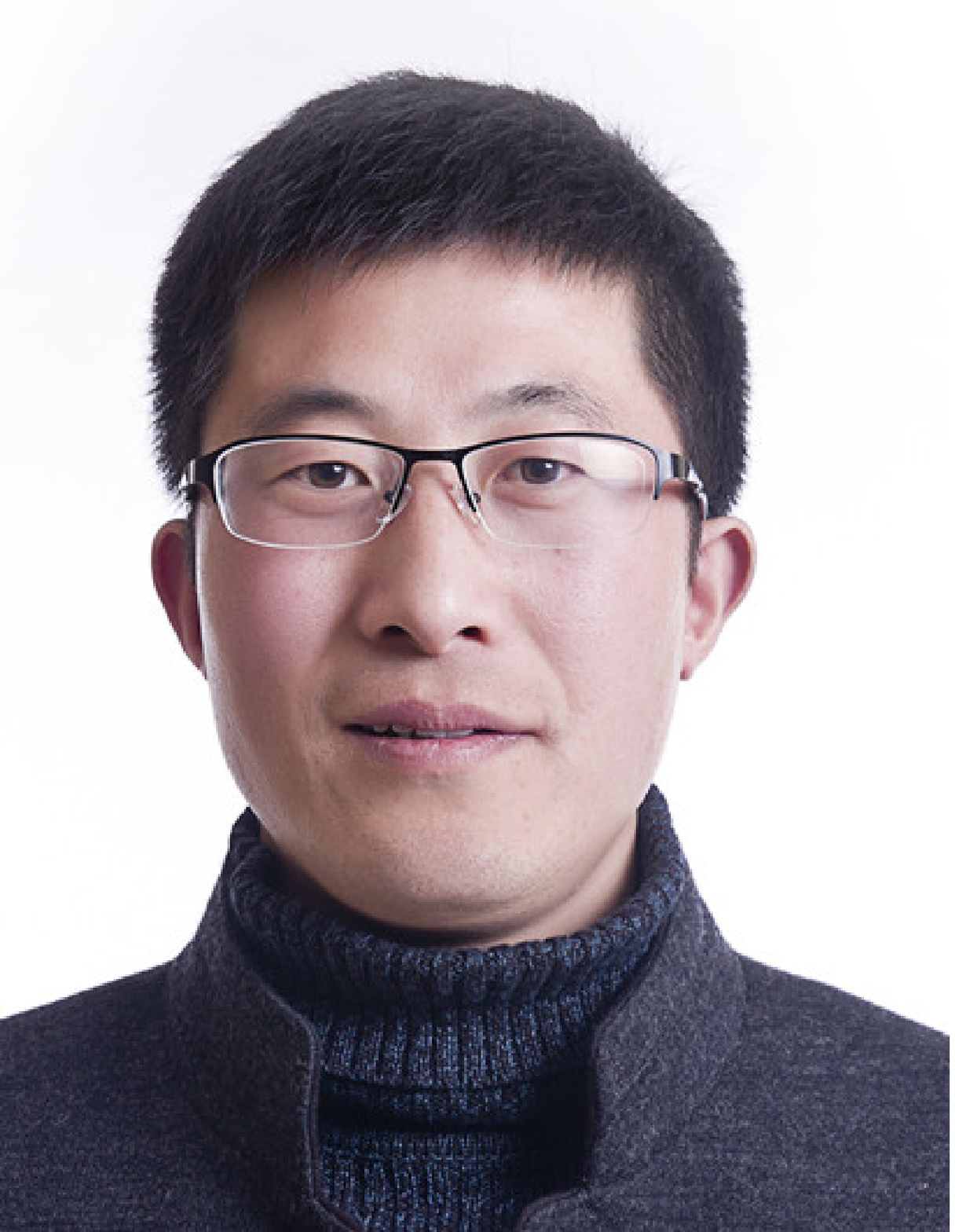}}]{Jielei Chu}
 received the B.S. degree from Southwest Jiaotong University, Chengdu, China in 2008, and is currently working toward the Ph.D. degree at Southwest Jiaotong University. His research interests are deep learning, data mining, semi-supervised learning and ensemble learning.
\end{IEEEbiography}
\begin{IEEEbiography}[{\includegraphics [width=1in,height=1.25in,clip,keepaspectratio]{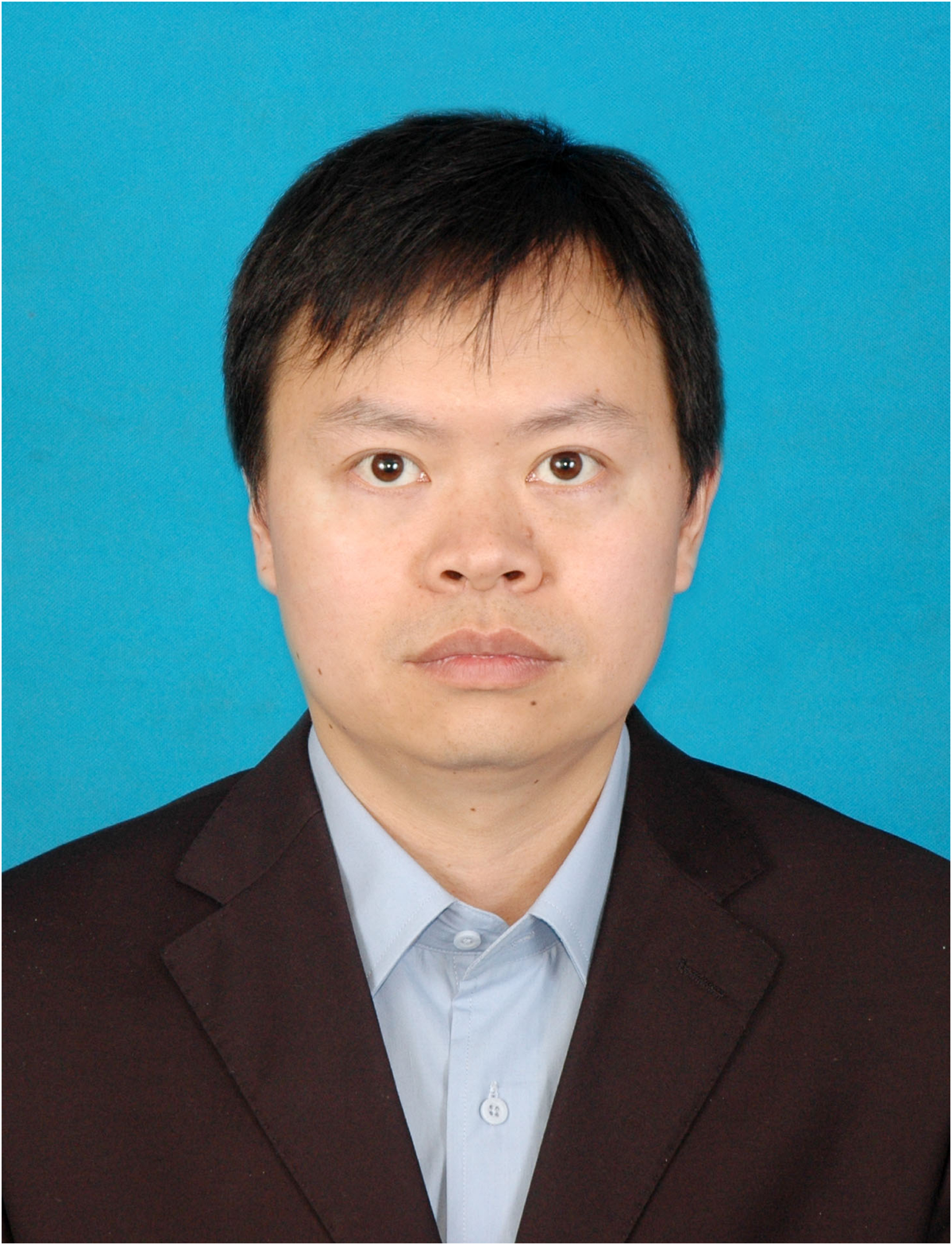}}]{Hongjun Wang}
 received his Ph.D. degree in computer science from Sichuan University of China in 2009.  He is currently Associate Professor of the Key Lab of Cloud Computing and Intelligent Techniques in Southwest Jiaotong University. His research interests are machine learning, data mining and ensemble learning. He published over 30 research papers in journals and conferences and he is a member of ACM and CCF. He has a reviewer for several academic journals.
\end{IEEEbiography}
\begin{IEEEbiography}[{\includegraphics [width=1in,height=1.25in,clip,keepaspectratio]{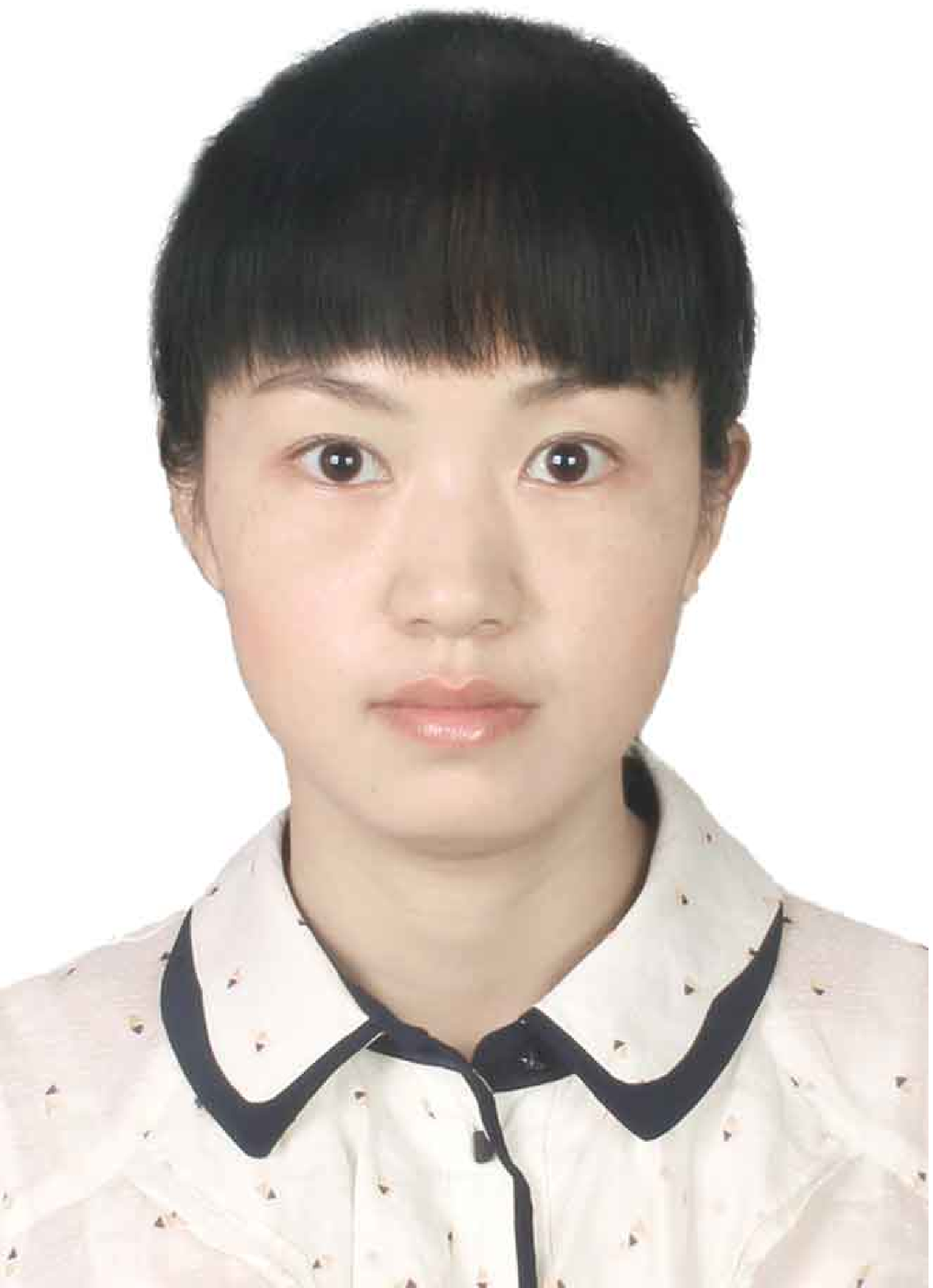}}]{Jing Liu}
 received his Ph.D. degree in management from Southwest Jiaotong University.She is currently an Assistant Professor of Business SchoolinSichuan University. Her research interestsare machine learning, financial technologyand modelling and forecasting high-frequency data.
\end{IEEEbiography}
\begin{IEEEbiography}[{\includegraphics[width=1in,height=1.25in,clip,keepaspectratio]{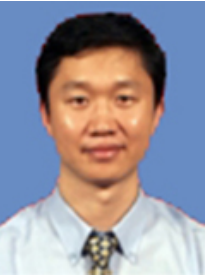}}]{Zhiguo Gong} received the Ph.D. degree in computer science from the Institute of Mathematics, Chinese Academy of Science, Beijing,China. He is currently a Professor with the Faculty of Science and Technology, University of Macau, Macau, China. His current research interests include machine learning, data mining, database, and information retrieval. \end{IEEEbiography}
\begin{IEEEbiography}[{\includegraphics [width=1in,height=1.25in,clip,keepaspectratio]{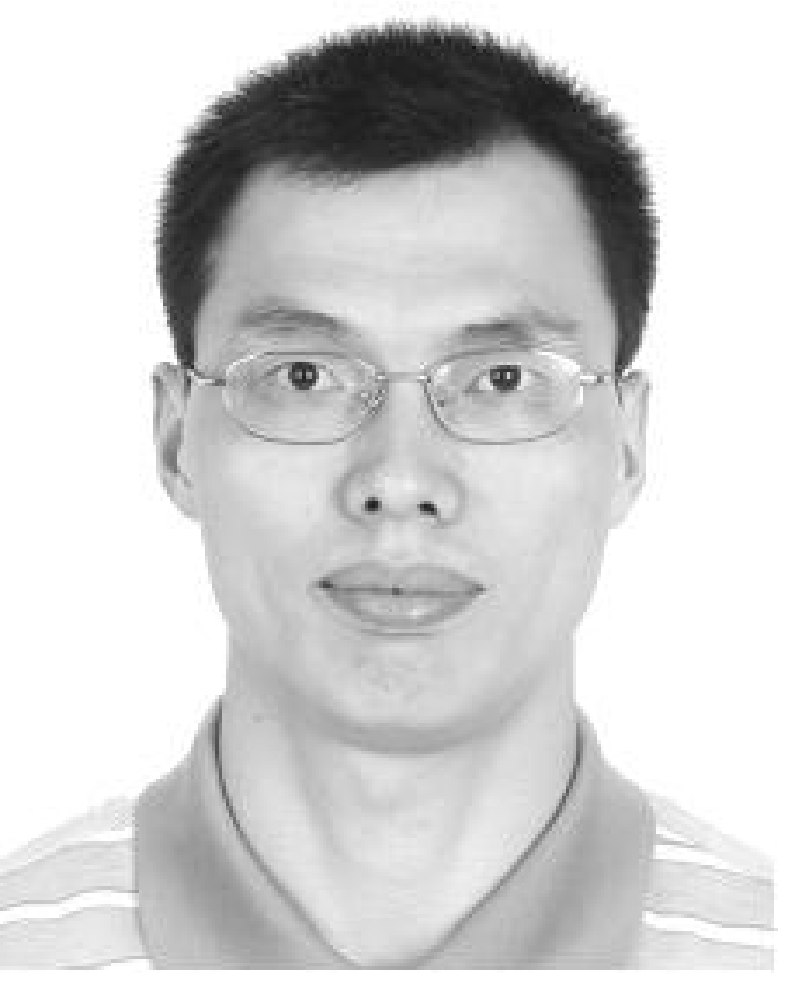}}]{Tianrui Li}
 (SM'11) received the B.S., M.S., and Ph.D. degrees in traffic information processing and control from Southwest Jiaotong University, Chengdu, China, in 1992, 1995, and 2002, respectively. He was a Post-Doctoral Researcher with Belgian Nuclear Research Centre, Mol, Belgium, from 2005 to 2006, and a Visiting Professor with Hasselt University, Hasselt, Belgium, in 2008; University of Technology, Sydney, Australia, in 2009; and University of Regina, Regina, Canada, in 2014. He is currently a Professor and the Director of the Key Laboratory of Cloud Computing and Intelligent Techniques, Southwest Jiaotong University. He has authored or co-authored over 150 research papers in refereed journals and conferences. His research interests include big data, cloud computing, data mining, granular computing, and rough sets. Dr. Li is a fellow of the International Rough Set Society.
\end{IEEEbiography}
\end{document}